\documentclass{article}

\usepackage[final]{nips_2017}
\usepackage[utf8]{inputenc} 
\usepackage[T1]{fontenc}    
\usepackage{hyperref}       
\hypersetup{
     colorlinks   = true,
     citecolor    = blue
}
\usepackage{booktabs}       
\usepackage{amsfonts}       
\usepackage{nicefrac}       

\usepackage{amsmath,amssymb}
\usepackage{float}
\usepackage[disable]{todonotes}
\usepackage{bm}

\newcommand{\R}{\mathbb{R}}
\newcommand{\Rp}{\mathbb{R}_+}
\newcommand{\bo}{\mathbf{1}}
\newcommand{\fU}{\mathcal{U}}
\newcommand{\fV}{\mathcal{V}}

\newcommand{\regional}{local}
\newcommand{\regions}{locales}
\newcommand{\regionalized}{localized}
\newcommand{\region}{locale}

\newcommand{\probability}{non-negative}

\newtheorem{defn}{Definition}

\usepackage{graphicx}

\title{Gini-regularized Optimal Transport with an Application to Spatio-Temporal Forecasting}
\author{Lucas Roberts \\
Amazon\\
\And Leo Razoumov\\ Amazon\\
\And Lin Su\\ NCSU \\
\And Yuyang Wang\\ Amazon\\}

\begin{document}

\maketitle

\begin{abstract}
Rapidly growing product lines and services require a finer-granularity forecast that considers geographic \regions. 
However the open question remains, how to assess the quality of a spatio-temporal forecast?
In this manuscript we introduce a metric to evaluate spatio-temporal forecasts. 
This metric is based on an Optimal Transport (OT) problem. 
The metric we propose is a constrained OT objective function using the \emph{Gini impurity function} as a regularizer. 
We demonstrate through computer experiments both the qualitative and the quantitative characteristics of the Gini regularized OT problem. 
Moreover, we show that the Gini regularized OT problem converges to the classical OT problem, when the Gini regularized problem is considered as a function of $\lambda$, the regularization parameter. 
The convergence to the classical OT solution is faster than the state-of-the-art Entropic-regularized OT\citep{cuturi2013sinkhorn} and results in a numerically more stable algorithm.
\end{abstract}


\section{Introduction}

Demand forecasting plays a central role in supply chain management, precipitating automatic ordering, in-stock management, and facilities location planning amongst others. 
 The classical forecasting problem in this context constitutes predicting sales, given a feature set, regardless the location of each sale. 
New product lines and services may exhibit rapid growth which requires a customized perspective in the forecasting domain. 
These challenges call for a \regionalized\ perspective which requires that we go beyond the coarser granularity and forecast the demand of a product for different \regions. 
Given a coarser granularity forecast for each product, one sensible approach to the spatio-temporal (\regional) forecast is to generate a forecast for each  \region, a partition of the coarser granularity forecast. 

\begin{figure}[H]
    \centering
    \begin{minipage}{.49\textwidth}
        \centering
        \includegraphics[width=\textwidth]{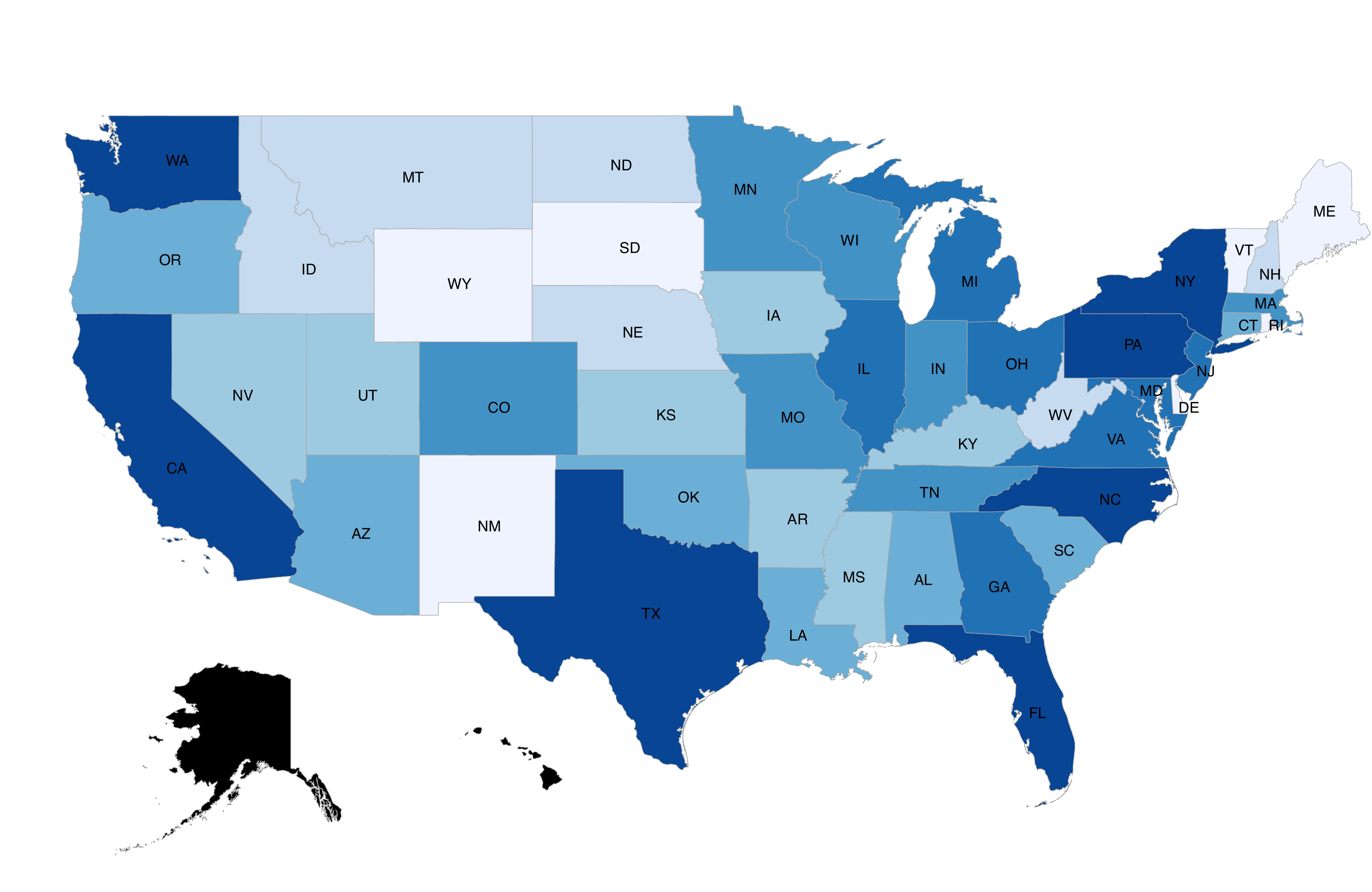}        
    \end{minipage}%
    \begin{minipage}{.49\textwidth}
        \centering
        \includegraphics[width=\textwidth]{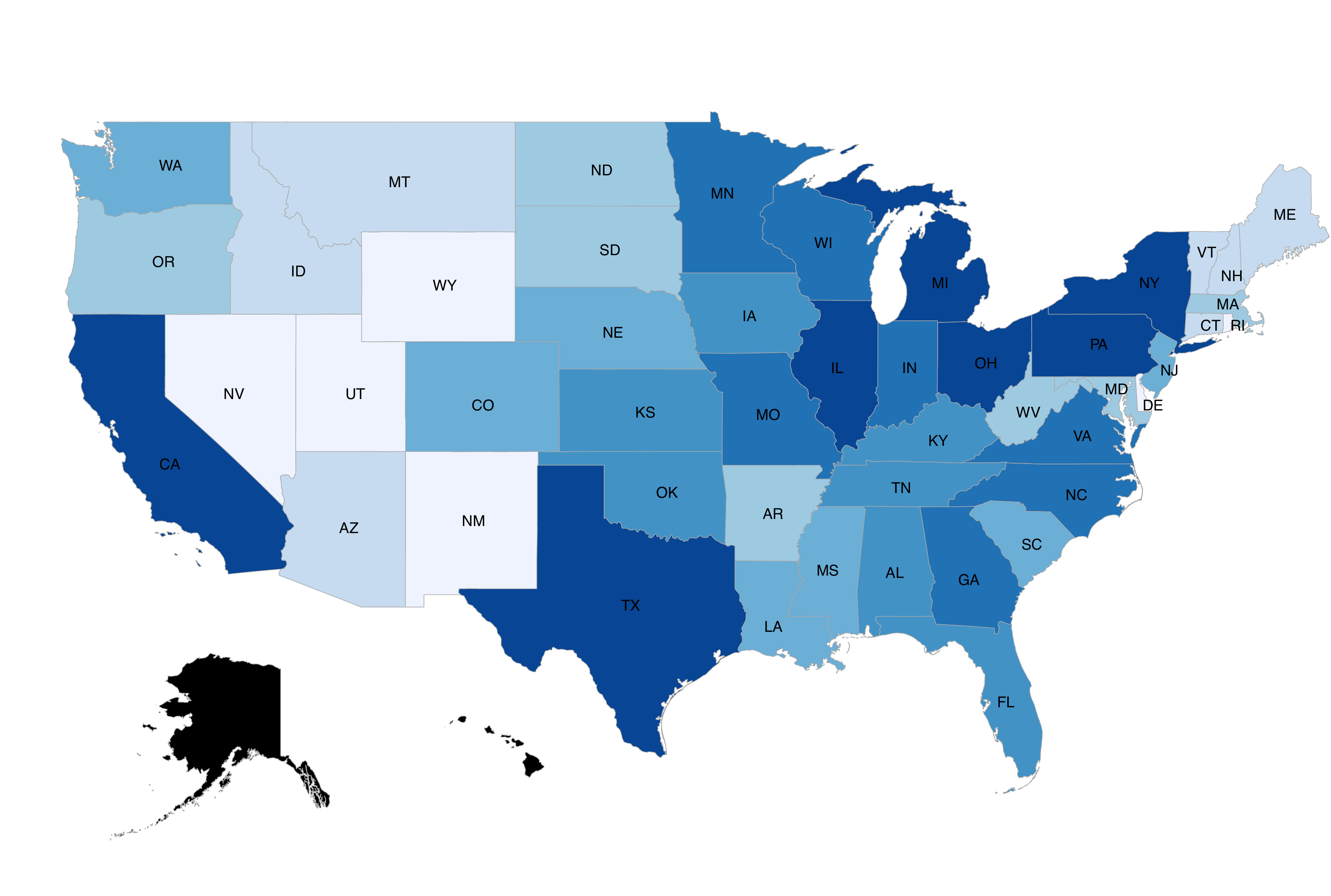}        
    \end{minipage}%
    \caption{Left: \regionalized\ forecasts Right:  \regional\ simulated sales. }
    \label{fig:pantry}
\end{figure}

Figure~\ref{fig:pantry} shows an example of \regional\ forecasting of simulated sales at the state level based on the population density of each state. 
\todo{remove sales}

Despite recent work on spatio-temporal (or \regional) 

assessing the quality of a forecast remains an open question. 
A desired metric should take the geometric aspects of the spatio-temporal forecast into consideration.
\todo[inline]{Leo: Unfortunately, I think we also need to make the paper devoid of any Amazon
context. I think we can still speak of forecasting but perhaps using a meteorological example/motivation
or perhaps another canonical example if anyone else can come up with one.}

This paper proposes a metric that is based on the \emph{optimal transport} distance as the forecasting accuracy criteria. 

Optimal Transport (OT), also known as the \emph{Wasserstein} or \emph{Earth Moving Distance} (EMD) provides a geographically congruous means to compare two \probability\ measures. 
Given the \probability\ measure $\bm{\mu}$, often called the source measure, and the \probability\ measure $\bm{\nu}$, often called the target measure, an OT plan measures the \emph{most frugal} way to transfer the \probability\ measure $\bm{\mu}$ to match the \probability\ measure $\bm{\nu}$ based on a \emph{ground metric}. 
Although the discrete OT problem may be cast as a linear program (LP), there is growing interest to approximate the classical OT solution with a regularized OT solution. 
One approach is Cuturi~\citep{cuturi2013sinkhorn}, who relaxes the original OT with an entropy regularizer.  
Adding the entropic regularizer makes the original OT problem \emph{strongly convex}, allowing for efficient computations and unique gradients \citep{frogner2015learning}. 
From an optimization perspective, adding a regularizer is similar to adding a barrier function. 
However, a major drawback of the entropic-regularized OT problem is that the solution algorithm requires choosing a large regularization parameter $\lambda$ to achieve a specified approximation error. 
A large value of the entropic regularization parameter leads to severe numerical issues (underflow/overflow) when using the Sinkhorn algorithm. 
The numeric instability of the entropic regularizer is documented in the literature with several modifications are proposed to stabilize the Sinkhorn methods~\citep{dvurechensky2017adaptive,chizat2016scaling,schmitzer2016stabilized}. 

\begin{figure}[!htb]
    \centering
    \begin{minipage}{.33\textwidth}
        \centering
        \includegraphics[width=\textwidth]{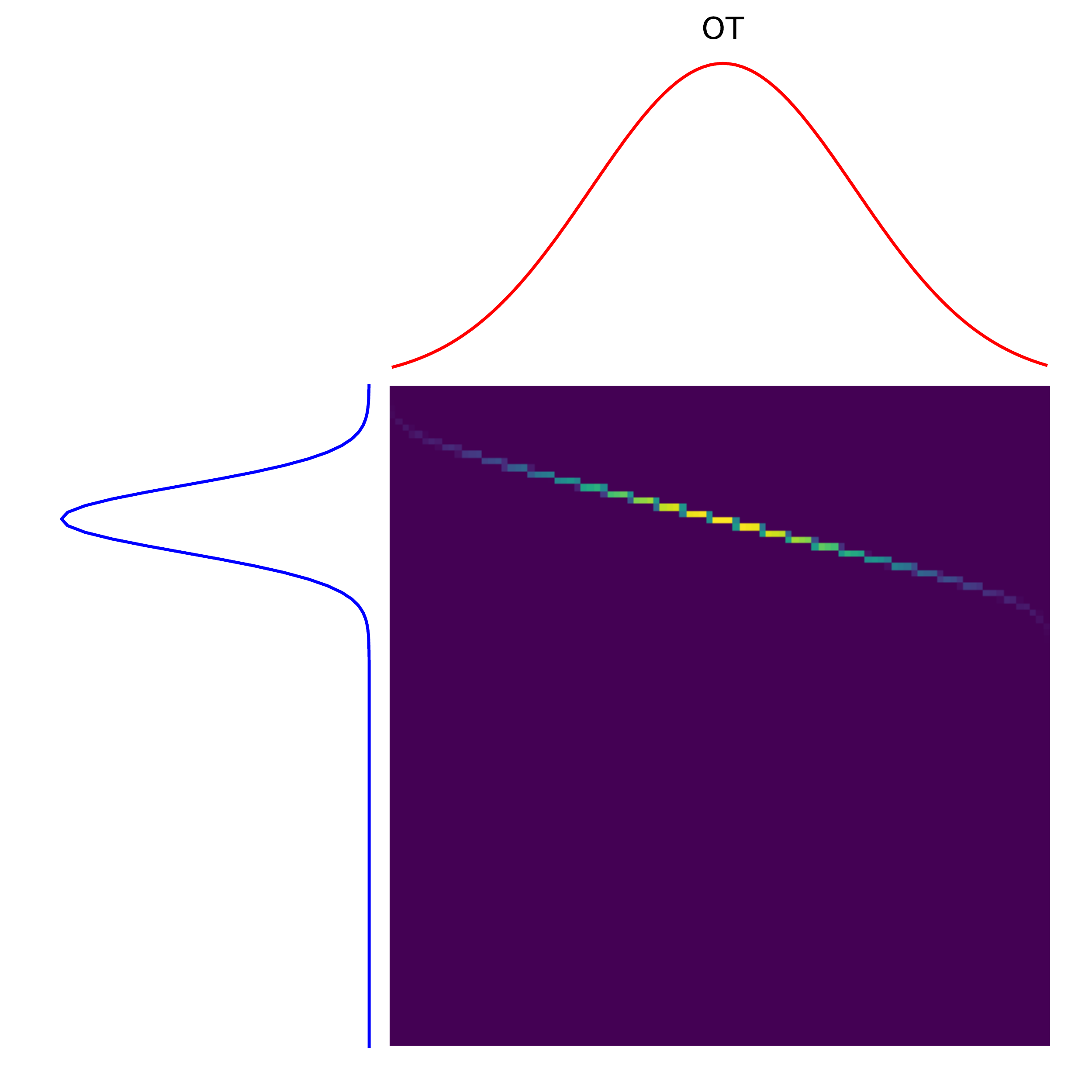}\\
        Optimal from LP        
    \end{minipage}%
    \begin{minipage}{0.33\textwidth}
        \centering
        \includegraphics[width=\textwidth]{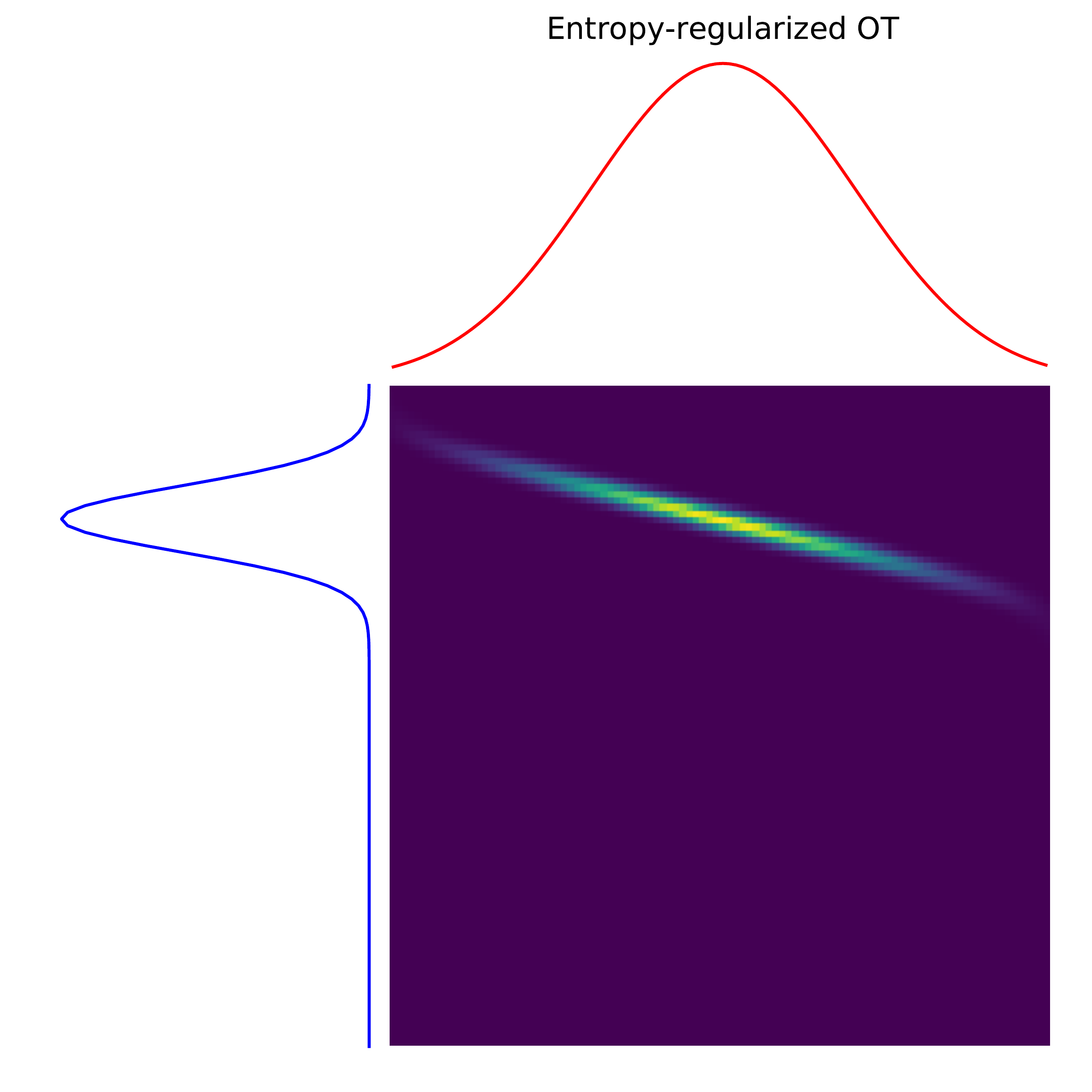}\\
          {EOT ($\lambda=1000$)}
    \end{minipage}
    \begin{minipage}{0.33\textwidth}
        \centering
        \includegraphics[width=\textwidth]{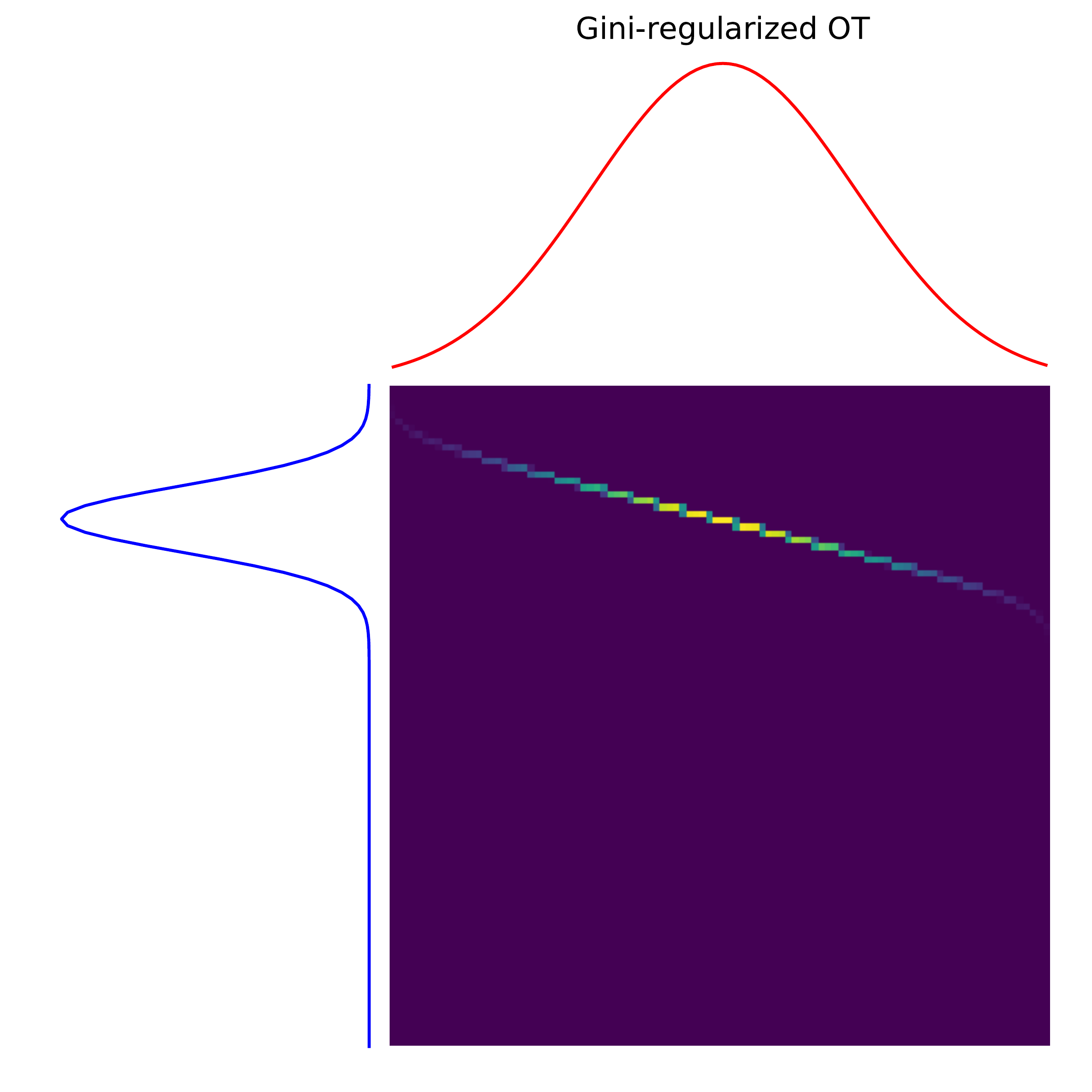}\\
        GOT ($\lambda=50$)        
    \end{minipage}
    \caption{The optimal transport plan between two 1D Gaussian (blue is the source and red is the target measure): (left) LP (as the ground truth); (center) Entropy-regularized OT; (right) Gini-regularized OT.}
    \label{fig:gauss1}
\end{figure}

\begin{figure}[!htb]
    \centering
    \begin{minipage}{0.24\textwidth}
        \centering
        \includegraphics[width=\textwidth]{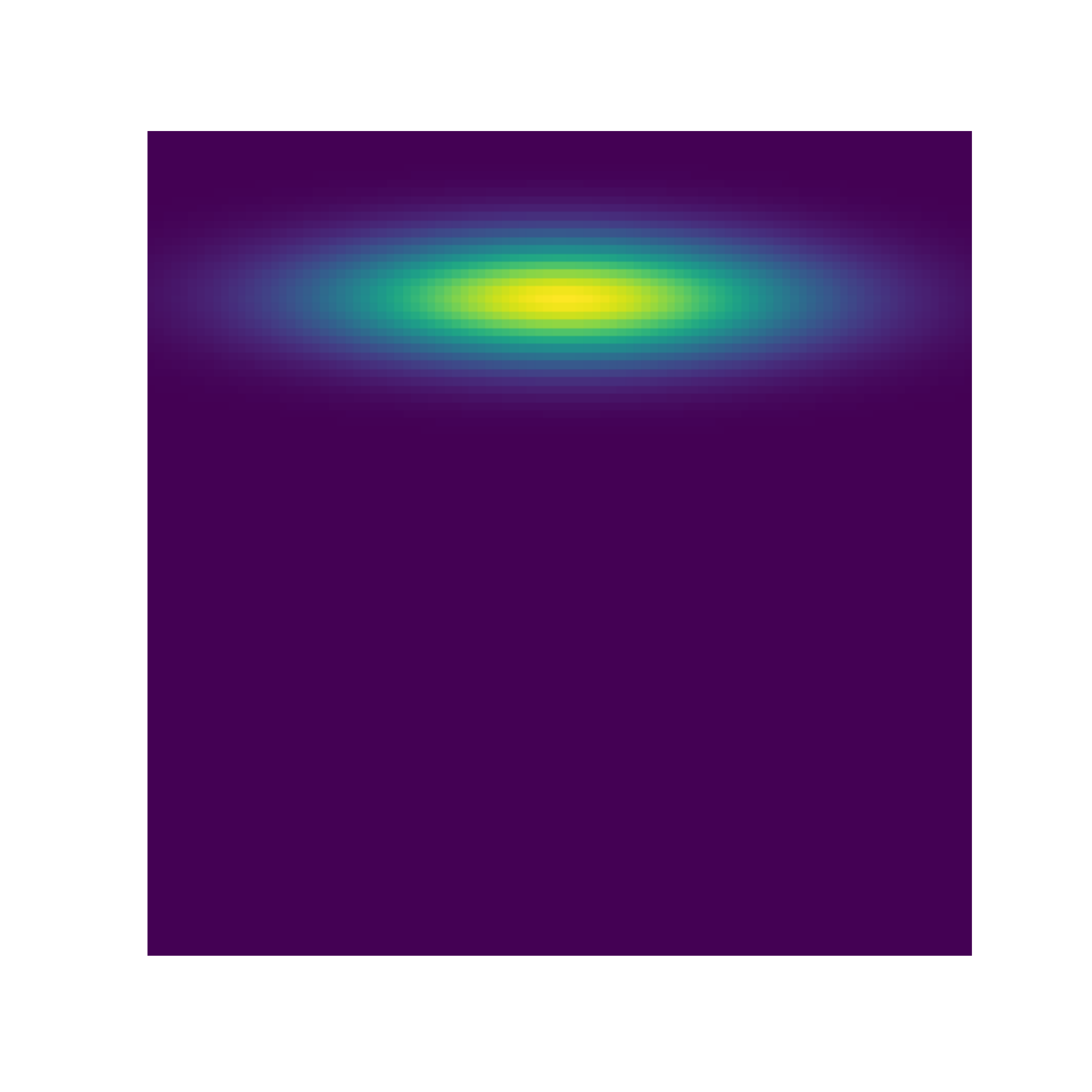}
    \end{minipage}
    \begin{minipage}{0.24\textwidth}
        \centering
        \includegraphics[width=\textwidth]{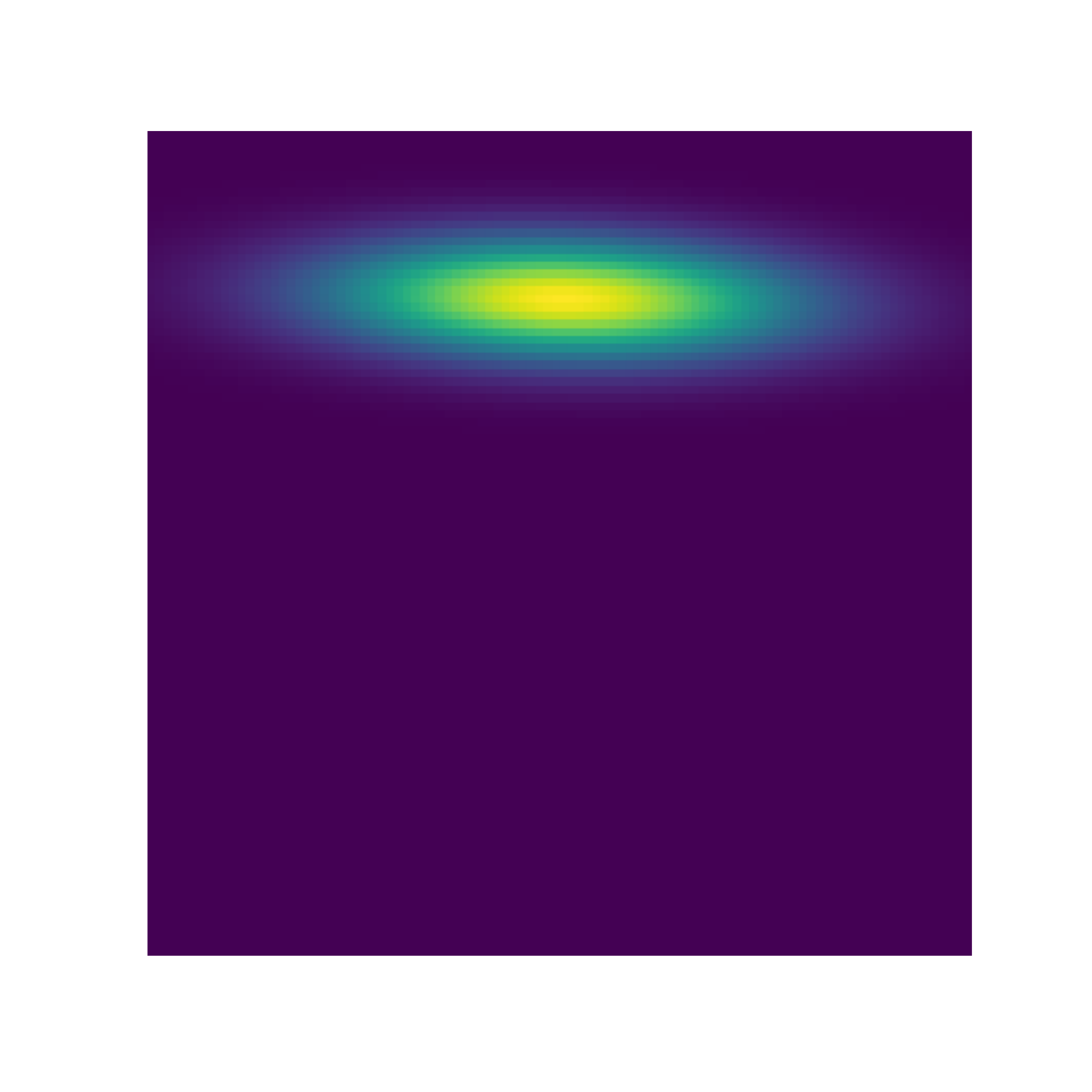}
    \end{minipage}
    \begin{minipage}{0.24\textwidth}
        \centering
        \includegraphics[width=\textwidth]{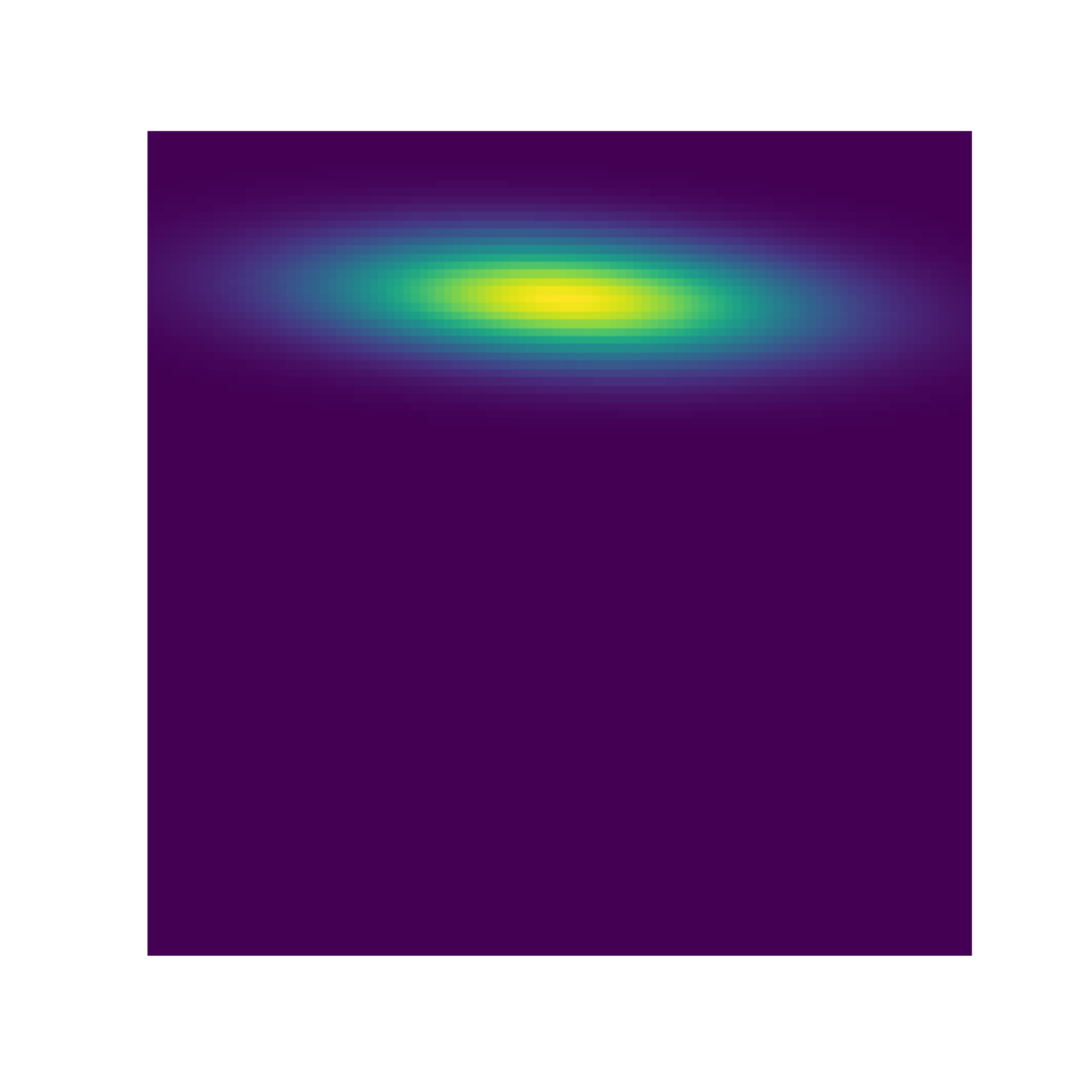}
    \end{minipage}
    \begin{minipage}{0.24\textwidth}
        \centering
        \includegraphics[width=\textwidth]{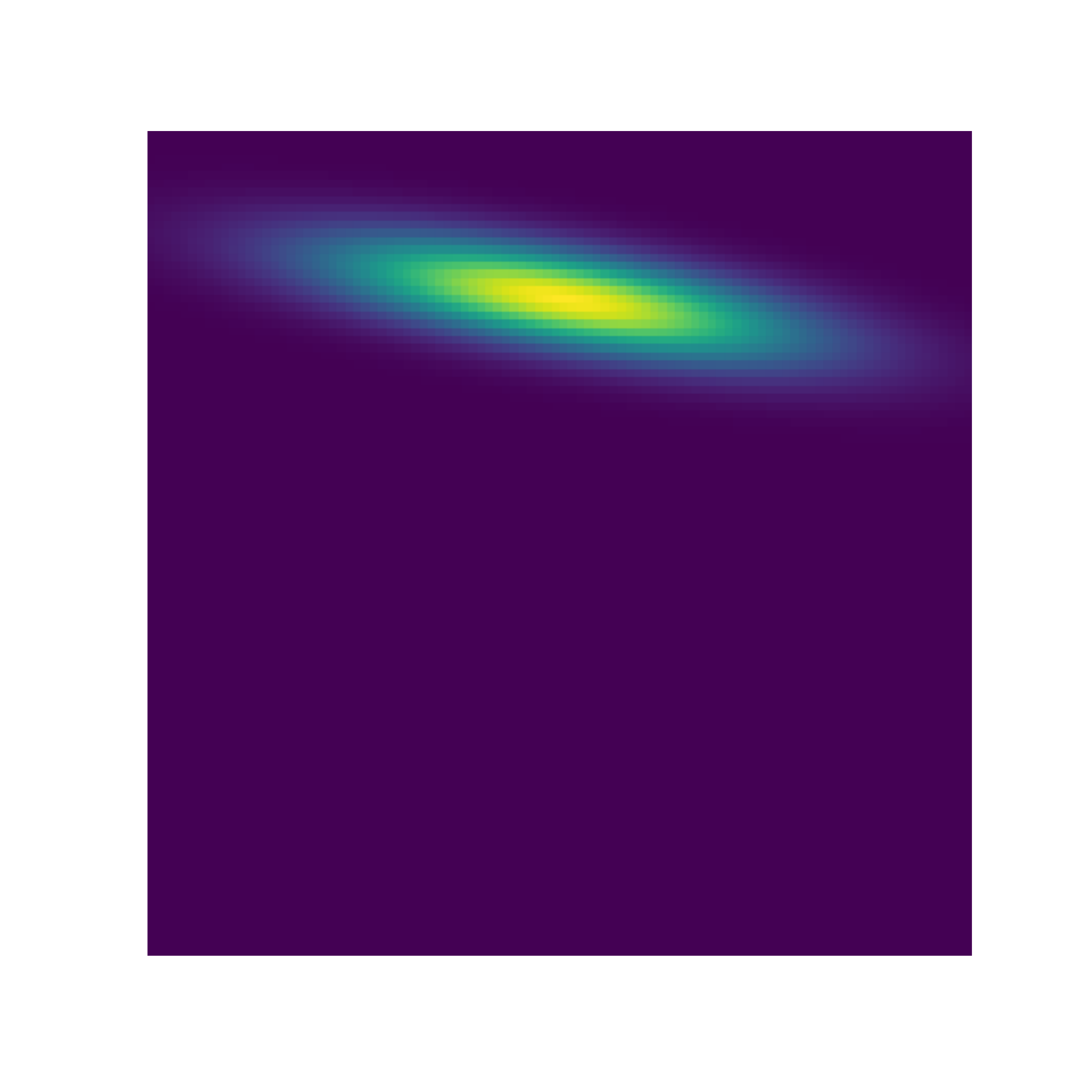}
    \end{minipage}
    \begin{minipage}{0.24\textwidth}
        \centering
        \includegraphics[width=\textwidth]{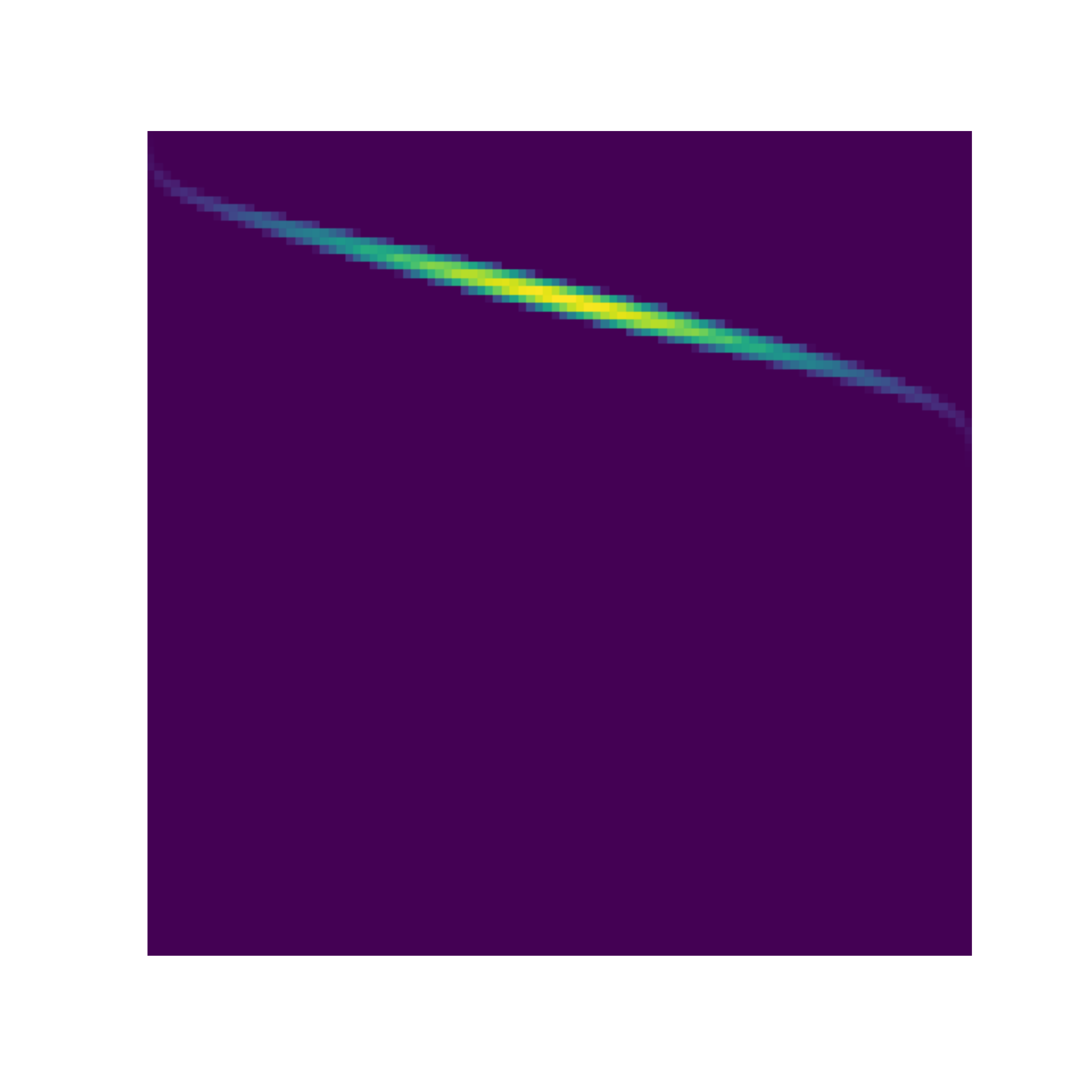}\\
        {$\lambda = 1$}
    \end{minipage}
    \begin{minipage}{0.24\textwidth}
        \centering
        \includegraphics[width=\textwidth]{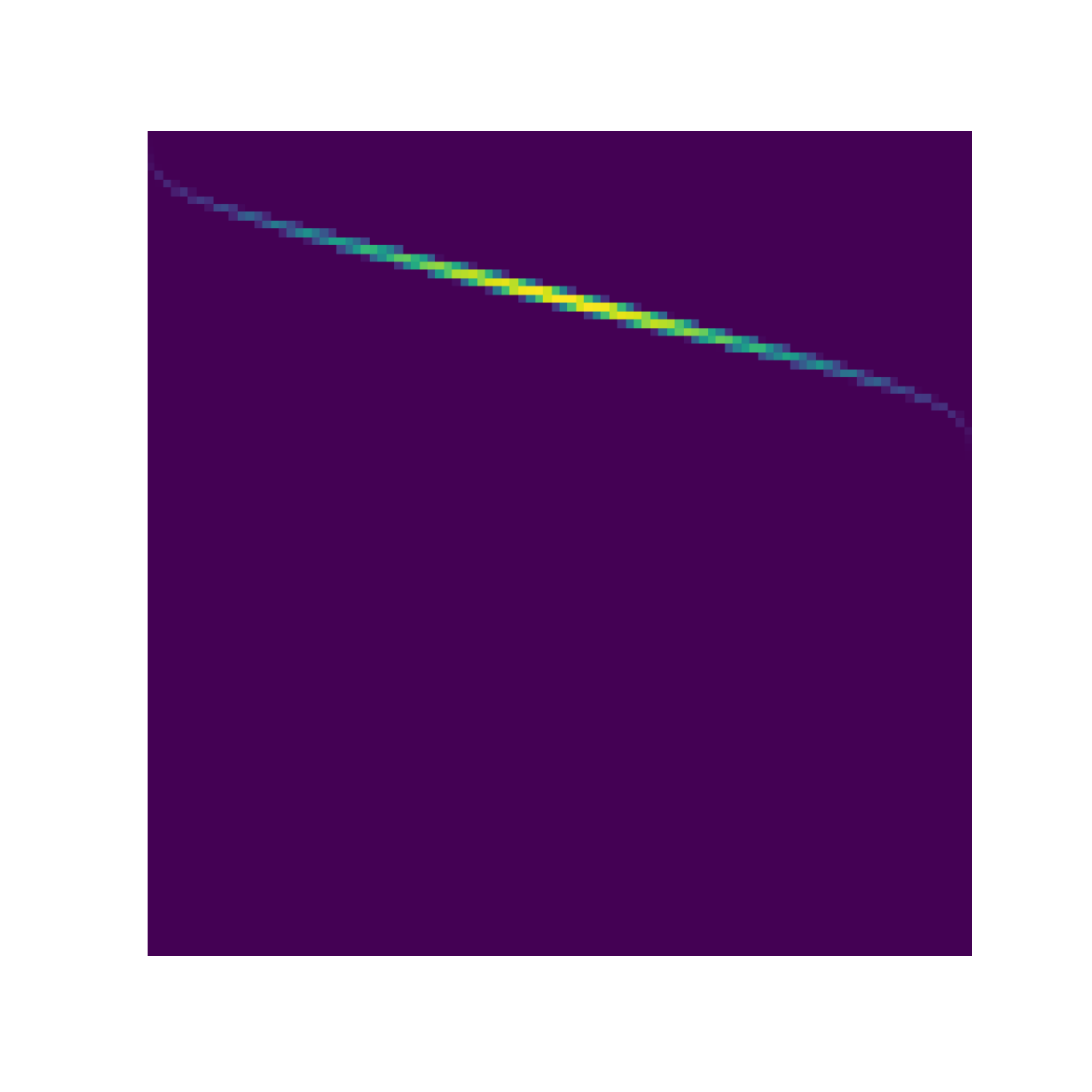}\\
        {$\lambda = 5$}
    \end{minipage}
    \begin{minipage}{0.24\textwidth}
        \centering
        \includegraphics[width=\textwidth]{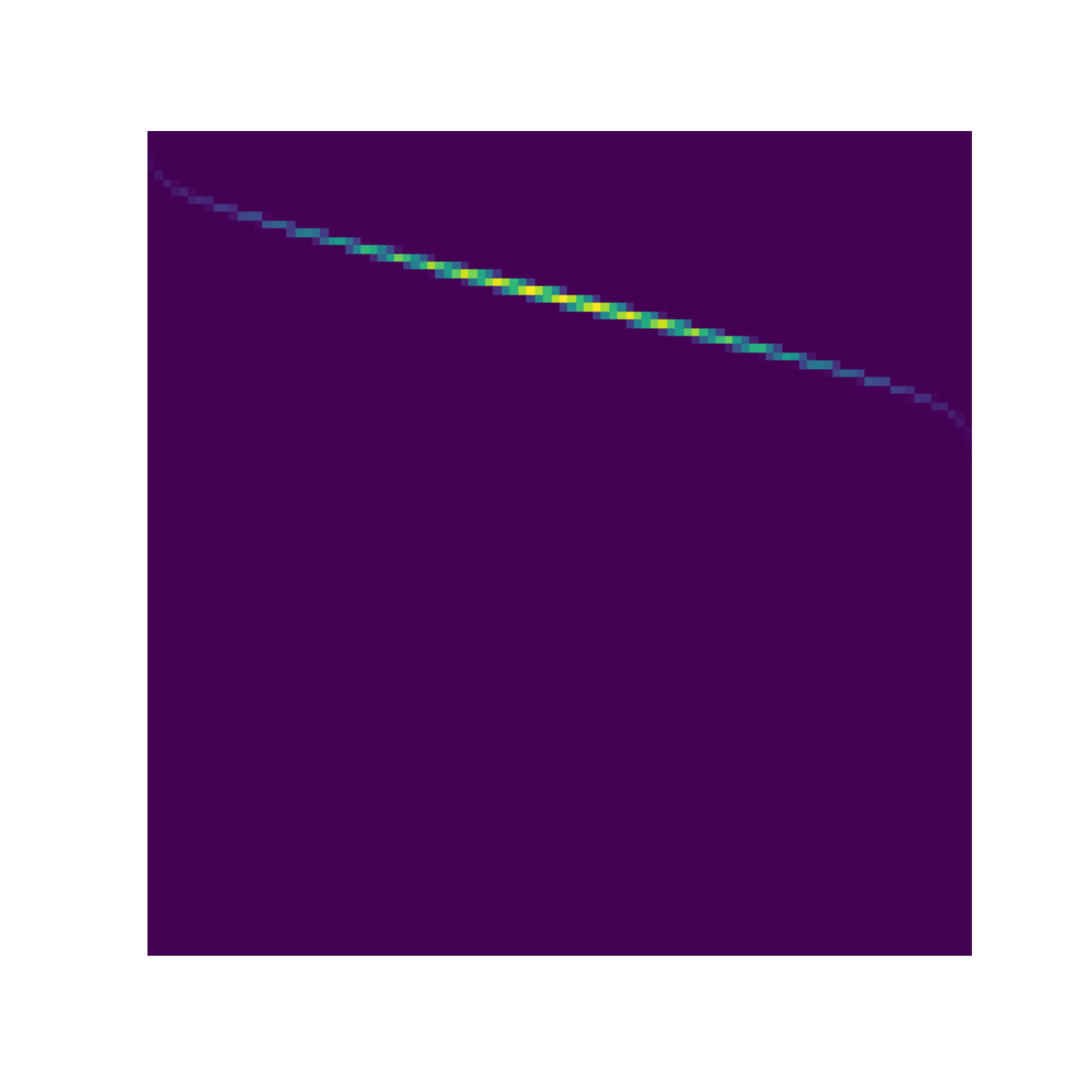}\\
        {$\lambda = 10$}
    \end{minipage}
    \begin{minipage}{0.24\textwidth}
        \centering
        \includegraphics[width=\textwidth]{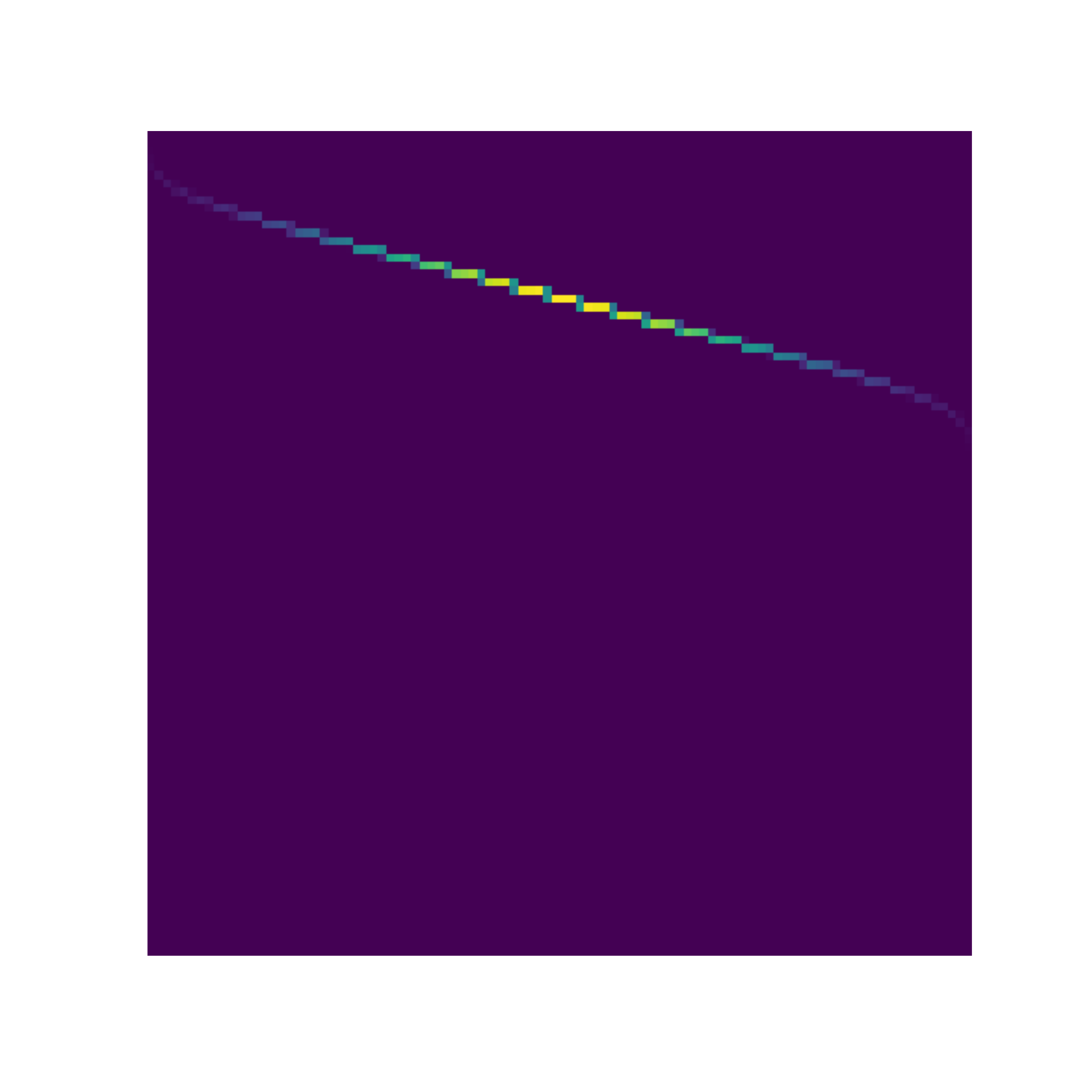}\\
        {$\lambda = 50$}
    \end{minipage}
    \caption{The optimal transport plan for entropy (top) and Gini (bottom) constraints with different regularization parameters.}
    \label{fig:gaussian2}
\end{figure}

In this paper, we study regularizing functions for OT problems and propose a quadratic regularizer, using the \emph{Gini impurity function} from the decision tree literature \citep{breiman1984classification}. 
A key advantage of the Gini regularizer is faster convergence than the entropy regularizer to the unconstrained OT solution when both are considered as a function of $\lambda$. 
Additionally, the Gini regularizer is less prone to numerical issues. 
Figure~\ref{fig:gauss1} shows that for the Gini regularizer, $\lambda=50$ produces better or at least as good results as the entropy constraints with $\lambda = 1000.$ 
Figure~\ref{fig:gaussian2} exemplifies the advantages of the convergence. 
Unlike the entropy regularizer, with a small $\lambda$ the Gini regularizer generates a near optimal OT plan. 
One might argue that the different scales of the two regularizers might lead to convergence differences, Figure~\ref{fig:gaussian2} shows otherwise with $\lambda = 50$ for entropy and $\lambda=5$ for Gini, both have the same scale but drastically different optimal transport plans.


Our contributions are twofold; firstly, we propose a forecasting accuracy criteria based on OT principles to evaluate spatio-temporal forecasts. To our knowledge, this is the first example of this application in the literature. 
Secondly, we propose a new OT regularizer which uses the Gini impurity function.  
We show through experiments the Gini regularizer is structurally dissimilar to the entropy regularizer for OT problems. 
We demonstrate empirically that when considered as a function of the regularization parameter $\lambda$, the Gini regularizer converges faster than the entropy regularizer to the LP solution.

The rest of the paper is organized as follows;
in Section~\ref{sec:bg}, we review the spatio-temporal forecasting problem, introduce the OT objective as the forecasting evaluation criteria, 
and introduce the Gini impurity function as an OT regularizer. 
In Section~\ref{sec:opt} we discuss the optimal transport problem and propose the Gini regularized OT problem along with the optimization algorithms, followed by numerical experiments in Section~\ref{sec:exp}. 
Finally, in Section~\ref{sec:conc} we conclude the paper with a summary of findings and point toward future works.

\textbf{Notations.} The probability simplex of dimension $K$ is denoted $\Sigma_K = \left\{\bm{\omega}\in\Rp^K: \sum_i\bm{\omega}_i = 1\right\}.$ 
We denote $\bo_K = [1, \cdots, 1]^T\in \mathbb{R}^K$ and drop the subscript $K$ whenever there is no confusion. 
The transport polytope is denoted $U(\bm{\mu}, \bm{\nu}) = \{\bm{P}\in\Rp^{n\times k}|\bm{P}\bo_k = \bm{\mu}, \bm{P}^T\bo_n = \bm{\nu}\}
$,  where $\bm{\mu}\in \Sigma_n$, and $\bm{\nu}\in \Sigma_k$.

 
\section{Background}
\label{sec:bg}

\subsection{EMD as an Accuracy Metric}

In this manuscript we use the word \region\ rather loosely, a \region\ might define states or provinces within a country, counties, interstate \regions\ (e.g. New England, Pacific Northwest etc.) or even cross national \regions\ (Scandinavia, Mediterranean, etc). 
Although making an accurate forecast in these situations is challenging, the earth moving distance we describe is broadly applicable in each of these scenarios.

However, a suitable metric to compare these models in a finer granularity is lacking. 
Current accuracy criteria such as the mean squared error (MSE) or the mean absolute percentage error (MAPE) cannot capture the relationship between forecasting errors  and the true cost of making an error. 
A requirement of any reasonable forecasting accuracy criteria is that the accuracy criteria reflect improvements in accuracy with improvements in the cost function. 

We propose to use the EMD as an accuracy criteria for spatio-temporal forecasting. 
The EMD was originally proposed as an OT distance between two \probability\ measures given a cost matrix $\boldsymbol{M}$. 
Conventionally, EMD is widely used as a measure of dissimilarity between two images in content-based image retrieval \citep{rubner2000earth, rubner1998metric}. 
We use the EMD as a spatio-temporal (\regional) forecast accuracy metric. 
In our case, the true demand distribution across \regions\ and the corresponding forecasting distribution across \regions\ are the two \probability\ measures of interest. 
EMD is a desired metric in situations where the entries of the cost matrix can be formulated by the researcher to reflect the true cost of an error in a decision based on the forecast. 

\subsection{Decision Tree Impurity Functions}
We provide a brief motivation for the investigation of the Gini regularizer and in the context of impurity functions from the decision tree learning literature. 

The motivation for investigating the Gini as a regularization function came by drawing an analogy between impurity functions that are used in the decision tree learning literature \citep{breiman1984classification} and the regularizing entropy proposed by Cuturi. By contrast, the Gini-regularized OT is a second-order polynomial objective function so that the solution method will be some variant of quadratic programming. 

Briefly, an impurity function is a convex downward function, that takes arguments in $p\in [0,1]$, the
arguments may be vector valued. Typical choices of impurity functions include: entropy, Gini, and
misclassification. We focus on the common impurity measure of Gini given the similarity with the
entropic regularizer, leaving the misclassification impurity to be investigated in future work.

\begin{figure}[!htb]
    \centering
    \begin{minipage}{.33\textwidth}
        \centering
        \includegraphics[width=\textwidth]{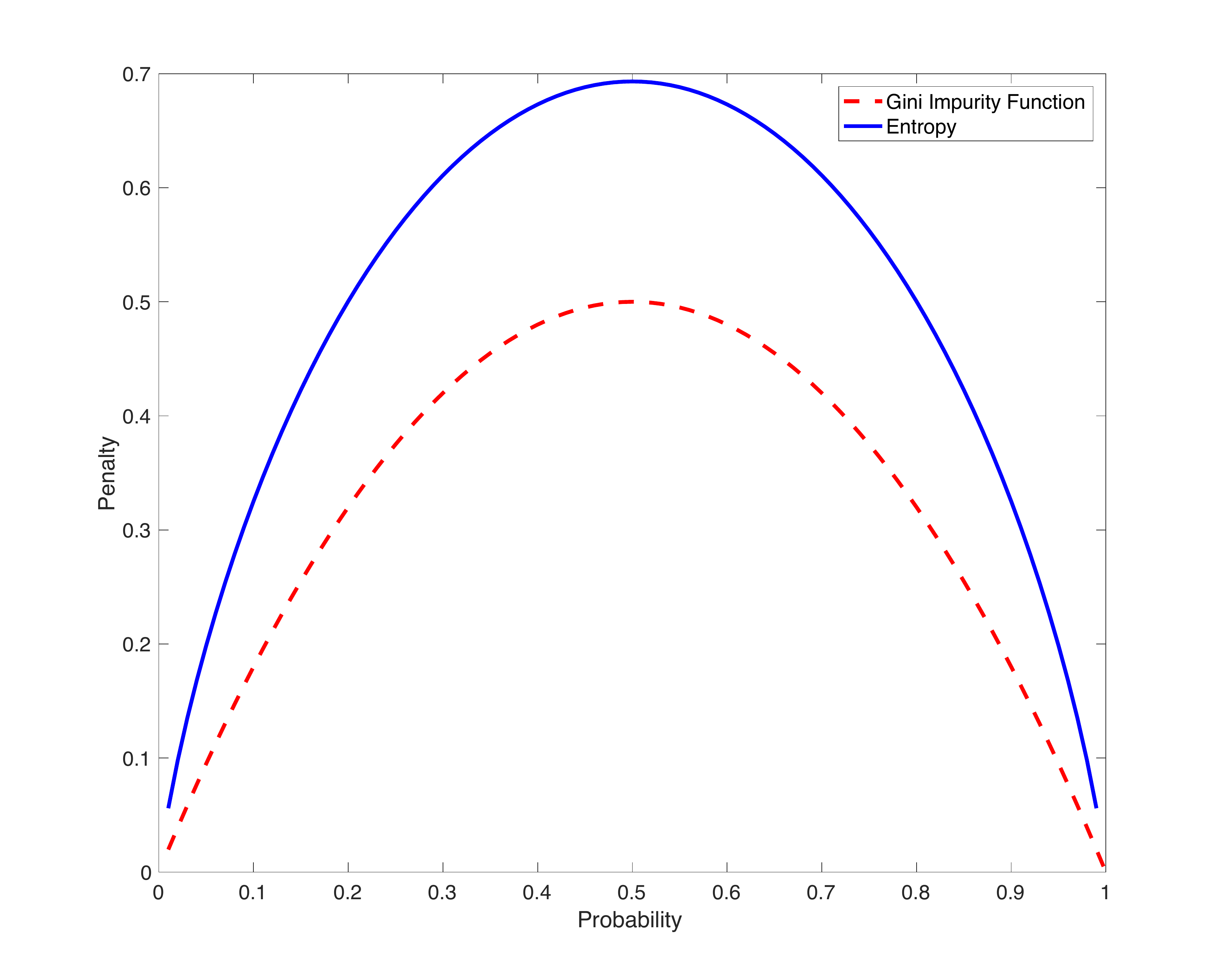}        
    \end{minipage}%
    \begin{minipage}{0.33\textwidth}
        \centering
        \includegraphics[width=\textwidth]{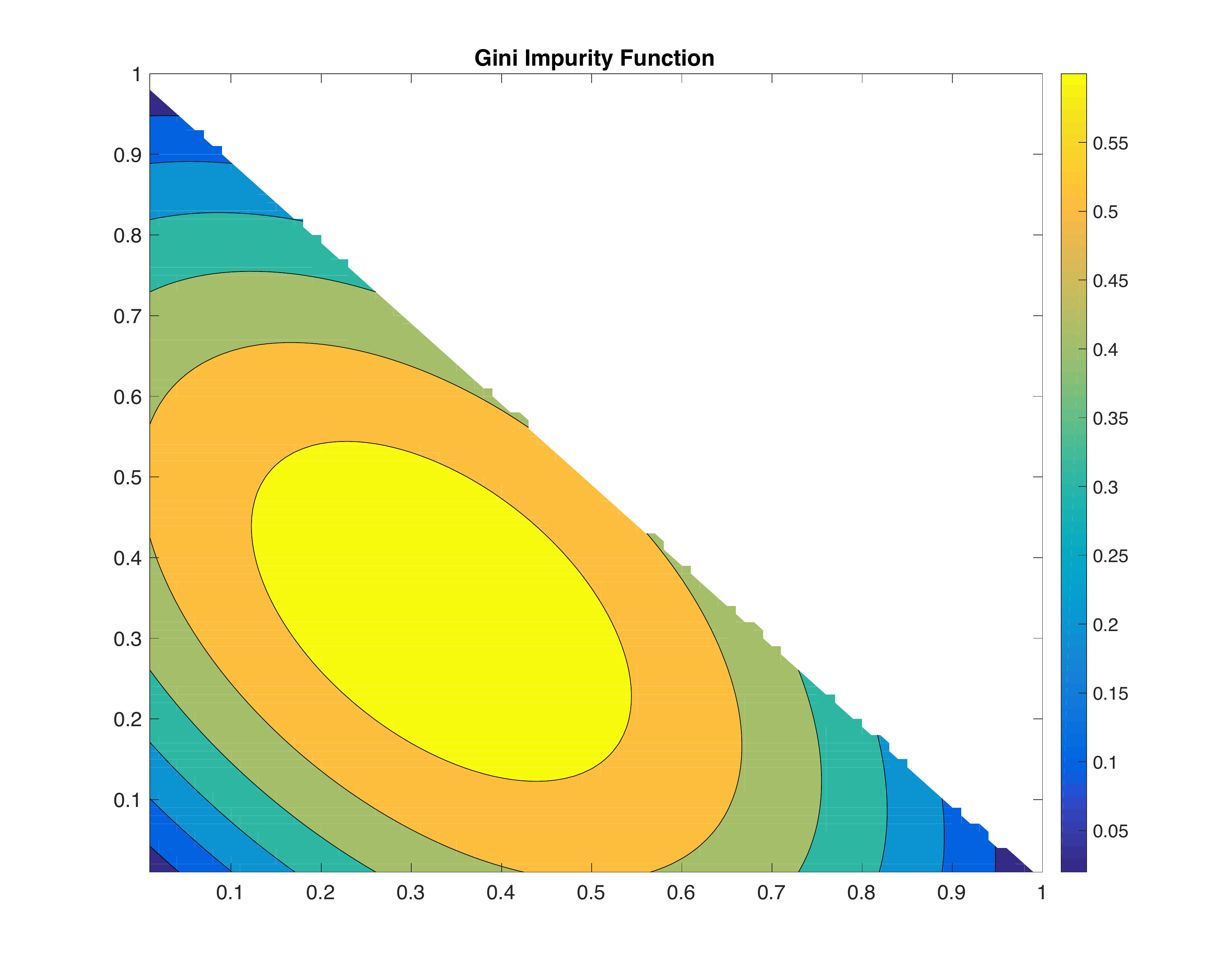}          
    \end{minipage}
    \begin{minipage}{0.33\textwidth}
        \centering
        \includegraphics[width=\textwidth]{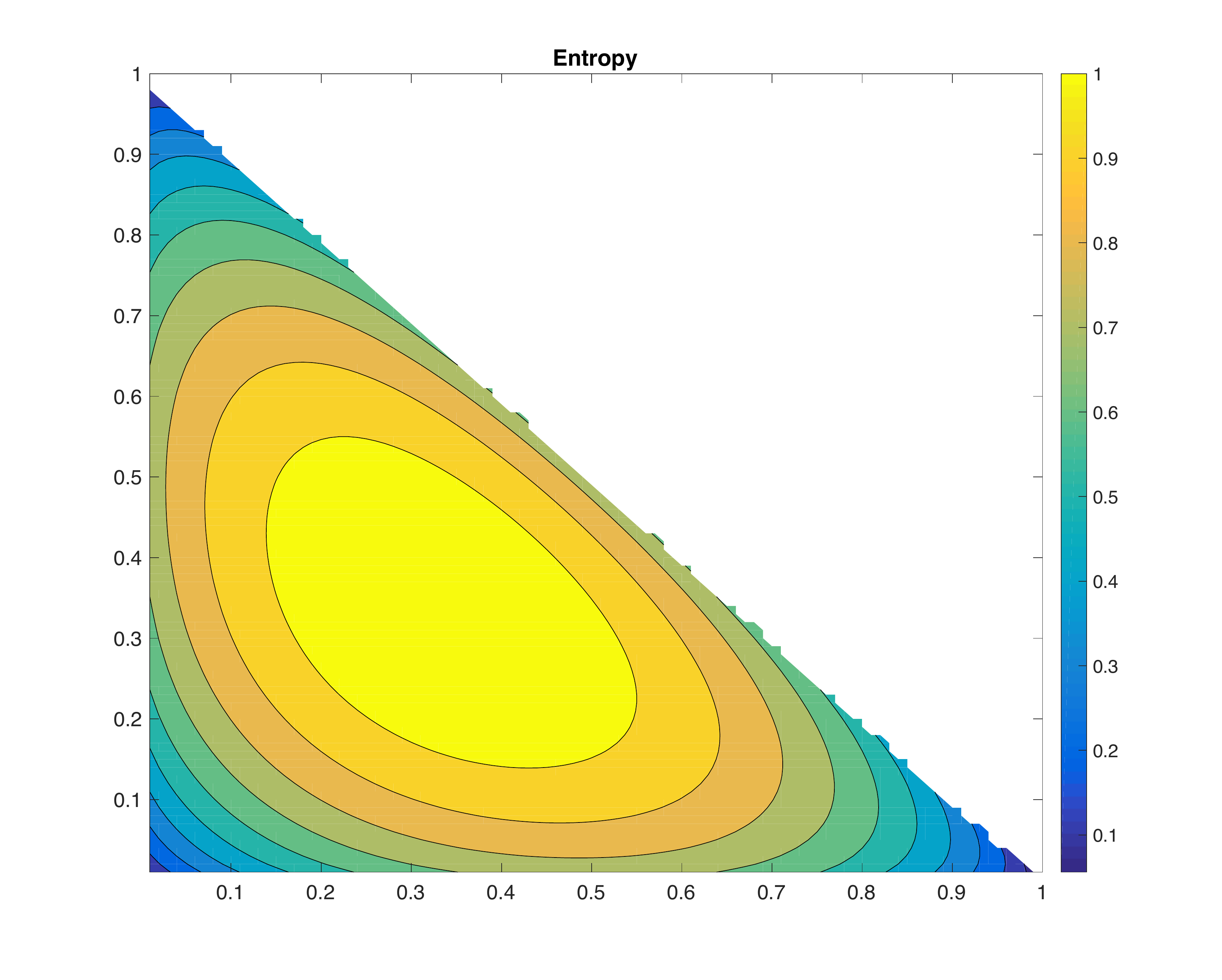}        
    \end{minipage}
    \caption{Magnitude of Entropy v.s. Gini when $k=2$ (left) and $k=3$ (middle and right)}
    \label{fig:entropy_gini_compare}
\end{figure}

Consider the graph in Figure \ref{fig:entropy_gini_compare} which gives both a qualitative and a quantitative description of both regularizing functions.
The graphs of both the entropy and the Gini regularizers look qualitatively similar, both accept probability arguments ($p\in [0,1]$), and both are defined as convex downwards functions.

The form of the regularizer induces useful properties in the OT problem.
Firstly, the penalty smoothes the kinked vertex points on the transportation 
polytope, yielding stable gradients \citep{frogner2015learning}. 
Secondly, the regularizer induces
structure in the optimization problem that allows for a tailored algorithm to be used
\citep{cuturi2013sinkhorn}. 
Finally, if the value of the OT problem at optimality is of interest,
the form of the regularizing function may facilitate better or worse approximations to the OT problem without
a regularizing term. This last item is the item of interest in this manuscript.  


\section{Optimal Transport and Regularized Variants}
\label{sec:opt}

In this section we first define the classical OT problem, then the entropy regularized OT problem, and finally the new Gini regularized OT problem.

\subsection{Classical OT and Entropy-regularized OT}
Let $\fU, \fV$ denote two metric spaces. 
Given a (continuous) cost function $c:\fU \times \fV \rightarrow \R$, the optimal transport of $\bm{\mu}\in\fU$ and $\bm{\nu}\in\fV$ is defined
\begin{equation}
\label{eqn:OP}
W_c(\bm{\mu}, \bm{\nu}) = \underset{\gamma\in \Pi(\bm{\mu}, \bm{\nu})}{\inf} \int_{\fU \times \fV} c(\bm{x}, \bm{y})\gamma({\rm{d}}\bm{ x}, {\rm{d}} \bm{y}),
\end{equation}
where $\Pi(\bm{\mu}, \bm{\nu})$ is the set of all \probability\ measures on the product space $\fU \times \fV$ such that $\bm{\mu}$ and $\bm{\nu}$ are the marginal measures on $\fU$ and $\fV$ respectively, often this representation is defined as a coupling space. 
The cost function $c(\cdot, \cdot)$-sometimes called the \emph{ground metric}-is a function that reflects the ``cost'' of moving a unit of mass from $\bm{x}$ to $\bm{y}$. 
The cost function $c(\cdot, \cdot)$ depends on the problem application and requires domain specific expertise to construct. 
When the spaces $\fU$ and $\fV$ coincide, a popular choice of cost function is the Euclidean metric (or its $p$-th power) and the resulting $W_c^p$ is called \emph{Wasserstein distance}. 
For example, for the Euclidean space, we have $c(\bm{s}, \bm{t}) = \|\bm{s} - \bm{t}\|^2_p.$

In the sequel, we consider the discrete OT problem, i.e., $\bm{\mu}$ and $\bm{\nu}$ are probability vectors in $\Sigma^K = \{\bm{\omega}\in{\Rp^K}: \mathbf{1}_K^T\bm{\omega} = 1\}.$ 
Given a cost matrix $\bm{M}\in \R^{n\times k}$, the optimal transport from the source measure $\bm{\mu}$ to target measure $\bm{\nu}$ is
\begin{equation}
\label{eqn:OPD}
W_{\bm{M}}(\bm{\mu}, \bm{\nu}) = \underset{\bm{P}\in U(\bm{\mu},\bm{ \nu})}{\min} \langle \bm{P}, \bm{M} \rangle, 
\end{equation}
where $\langle \bm{P}, \bm{M}\rangle = \textit{trace}(\bm{P}^T\bm{M})$ and 
$U(\bm{\mu, \nu}) = \{\bm{P}\in\Rp^{n\times k}|\bm{P}\bo_n = \bm{\mu}, \bm{P}^T\bo_k = \bm{\nu}\}$. 
Note that when $\boldsymbol{M}$ itself is a metric matrix, EMD is a true distance satisfying all the distance axioms (non-negativity, identity of indiscernables, symmetry, and triangle inequality)  simultaneously. 
In this paper, we address the discussion in a more general framework, where $\boldsymbol{M}$ is not restricted to a metric matrix. 
EMD remains a valid cost measurement and all computational derivation is the same. 
We use the word "distance" rather loosely and EMD is not necessarily a mathematical distance in our general formulation. 

The classical OT problem on a finite space is an LP problem.
\begin{defn}
	Given an $n \times k$ cost matrix $\bm{M}$, the optimal transport problem is 
	\begin{align}
		\label{eqn:lp}
		\underset{\bm{P}\in U(\bm{\mu, \nu})}{\min} \langle \bm{P}, \bm{M} \rangle .
	\end{align}
\end{defn}
The resulting optimal matrix $\bm{P}^*$-the argument at optimality-is called a
flow matrix or optimal transport plan. 
The corresponding minimum distance $\langle \bm{P}^*, \bm{M} \rangle$ is denoted $\rho^*$.

%

The exact evaluation of the OT loss $W_{\bm{M}}$ requires solving a linear program that is not only computationally costly but also leads to an unstable gradient \citep{frogner2015learning}. 
In his seminal paper, \cite{cuturi2013sinkhorn} proposes a relaxed or smoothed version of the OT problem by adding an entropic constraint that induces computational advantages.
\begin{defn}
	Given an $n \times k$ cost matrix $\bm{M}$, the entropy-regularized OT (EOT) problem is 
	\begin{align}
		\label{eqn:eot}
		\underset{\bm{P}\in U(\bm{\mu, \nu})}{\min} \langle \bm{P}, \bm{M} \rangle - \frac{1}{\lambda}h(\bm{P}), 
	\end{align}
	where $h(\bm{P})=-\sum_{i=1}^{n}\sum_{j=1}^{k}P_{ij}\log P_{ij}$ is the entropy-regularizer. Note here the convention is $0log(0)=0$.
\end{defn}
The entropy-regularized EMD has a nice interpretation as the Kullback-Leibler divergence and can be solved
by applying Sinkhorn's theorem. 
Specifically, \cite{cuturi2013sinkhorn} shows that the optimal
transport matrix $\bm{P}^*_e(\lambda)$ that solves Equation \eqref{eqn:eot} is a rescaling of matrix
$\exp(-\lambda\cdot \bm{M})$, where $\exp(\cdot)$ is the element-wise exponential. 
The solution can be found by the Sinkhorn algorithm. 
The notation $\bm{P}^*_e(\lambda)$ and $\rho_e(\lambda)=\langle \bm{P}^*_e(\lambda),
\bm{M} \rangle$ are used to represent the optimal entropy-regularized solution and the distance respectively.

Here we present a simple \emph{information theoretic} motivation for the KL based regularized
Optimal Transport. 
As an illustration we consider moving grain from multiple granaries with
normalized capacity described by a discrete probability distribution $\bm{\mu}$ to multiple bakeries
with normalized demand described by a discrete distribution $\bm{\nu}$. 
For a given (feasible) transportation plan $\bm{P}$ the pure transportation costs are given by $\langle \bm{P},\bm{M} \rangle$. 
Every grain
transporter needs to be communicated at least 1 source-destination location tuples denoted $(i,j)$ in order to
operate. 
If the cost of such communication is non-negligible it has to be counted into the overall
transportation cost as well.  
A minimal number of \emph{extra} bits needed to encode a pair $(i,j)$
distributed with $\bm{P}$ using a codebook matched to default distribution $\bm{\mu}\otimes\bm{\nu}$ is
given by the $\mathrm{KL}(\bm{P}\|\bm{\mu}\otimes\bm{\nu})$ \citep{cover2006,gruenwald2007}. 
Assuming $\lambda$ bits/currency unit, the corresponding contribution to the overall transportation plus communication costs would be exactly $\frac{1}{\lambda}\mathrm{KL}(\bm{P}\|\bm{\mu}\otimes\bm{\nu})$ --the KL regularizer term.  
Similar information theory considerations can be applied to the entropic regularizer as well.

Despite the appealing acceleration of EOT compared to an LP solution, the Sinkhorn algorithm \citep{sinkhorn1967diagonal} can exhibit severe numerical issues. 
The assumption in the Sinkhorn algorithm is that the matrix  $\boldsymbol{A}=\exp(-\lambda \boldsymbol{M})$ for EOT is strictly positive. 
Although in theory $\exp(\cdot)$ is always positive, when $\lambda$ becomes large, in any finite state machine the  $\boldsymbol{A}_{ij} \to 0$. 
Consider the case when $\boldsymbol{M}$ has nonzero entries in the off-diagonals, as $\lambda \to \infty$ overflow or underflow issues will appear after several iterations. 
In this case the Sinkhorn iteration cannot stop if $\boldsymbol{u} \neq \boldsymbol{v}$ until overflow or underflow happens. 
On one hand, the parameter $\lambda$ must be large for the solution to the EOT problem approximates the classical OT problem. 
On the other hand, the Sinkhorn algorithm has numerical issues when $\lambda$ increases. 
Therefore, a stable approach that converges faster than the EOT to the classical OT as a function of $\lambda$ is necessary.

\subsection{Gini-regularized Optimal Transport}
We define the Gini-regularized OT (GOT) problem and the corresponding \textsc{got} distance denoted $\rho_g(\lambda)$ as follows:
\begin{defn}
Given an $n \times k$ cost matrix $\bm{M}$, the Gini-regularized OT (GOT) problem can be written
\begin{equation}
\label{eqn:got}
\underset{\bm{P}\in U(\bm{\mu, \nu})}{\min} \langle \bm{P}, \bm{M} \rangle - \frac{1}{\lambda}\mathcal{G}(\bm{P}). 
\end{equation}
 Define the Gini regularizer 
\[\mathcal{G}(\bm{P}) = \sum_{i=1}^{n}\sum_{j=1}^k P_{ij}(1 - P_{ij}) = \textit{trace}\left[ \bm{P}^T(\bm{1}_n\bm{1}_k^T - \bm{P})\right]  = \langle \bm{P}, \bm{1}_n\bm{1}_k^T - \bm{P}\rangle.\] 
We denote the resulting distance $\rho_g(\lambda) = \langle \bm{P}^*_g(\lambda), \bm{M} \rangle$ the
\textsc{got} distance with $\bm{P}^*_g(\lambda)$ being the argument at the optimal solution of Equation \eqref{eqn:got}.
\end{defn}

Let us look closer at the objective function in Equation \eqref{eqn:got}, which can be re-written as
\begin{equation}
\label{eqn:got2}
\underset{\bm{P}\in U(\bm{\mu, \nu})}{\min} \langle \bm{P}, \bm{M}_\lambda \rangle + \frac{1}{\lambda} \|\bm{P}\|_F^2 \stackrel{.}{=} f(\bm{P}), 
\end{equation}
where $\bm{M}_\lambda = \bm{M} - \bm{1}_n\bm{1}_k^T / \lambda.$ and $\|\cdot\|_F$ denotes the Frobenius norm.

We explore three different methods to solve Equation \eqref{eqn:got2}. 
First, one may show that Equation \eqref{eqn:got2} is a standard quadratic programming problem with respect to $\textnormal{vec}(\bm{P})$. 
Thus, Equation \eqref{eqn:got2} can be solved efficiently by any generic quadratic programming solver. 
We use a python-embedded package called CVXPY \citep{cvxpy}, a modeling interface for convex optimization. 
CVXPY relies on the Embedded Conic Solver (ECOS) \citep{bib:Domahidi2013ecos} to solve classes of conic optimization problems such as quadratic programs. 
The second method is due to~\citep{ferradans2014regularized} who proposes the conditional gradient (or Frank-Wolfe) method to solve the regularized OT problem. 

Finally, it is immediately verified that Equation \eqref{eqn:got2} is strongly convex and
smooth, for which \emph{mirror descent} (MD) 
\eqref{eqn:got2}~\citep[see][Section 4.3]{bubeck2015convex} is well suited. 
In a nutshell, MD works in a similar
fashion as \emph{projected gradient descent}. 
A point is first mapped (via the so-called \emph{mirror
map}) to the dual space, and one takes a gradient step and reverses the mirror map to get the candidate
point for the next iteration in the primal space. 
Since the candidate point may fall outside of the
feasible region, one needs to take a projection with a \emph{Bregman divergence.} 
Specifically for the simplex setup, the canonical mirror map is given by the negative entropy~\citep[see][Section 4.3]{bubeck2015convex}, which yields the following updates,
\[
\begin{split}
\widetilde{\bm{P}}_{t+1}  &= \bm{P}_{t}\odot \exp(-\eta\cdot\nabla f(\bm{P}_{t})), \\
\bm{P}_{t+1} &\in \Pi_{U(\bm{\mu, \nu})} (\widetilde{\bm{P}}_{t+1}), 
\end{split}
\]
where $\eta$ is the step size and $\Pi_{U(\bm{\mu, \nu})}(\cdot)$ is the projection operator onto the polytope $U(\bm{\mu, \nu}).$ \cite{benamou2015iterative} shows that the projection step can be achieved by iteratively normalizing the rows and columns of $\widetilde{\bm{P}}_{t+1}$ so that $\widetilde{\bm{P}}_{t+1}$ have the desired row and column sums~\citep[see][Proposition 1]{benamou2015iterative}.

\cite{muzellec2017tsallis} applied MD to solve optimal transport problems. 
They generalized the entropy regularizer to a Tsallis regularizer. 
In their work, a general form of penalty-indexed by a single parameter $q$-is provided that covers many existing optimal transport regularizers as specific values of $q$. 
However, performance of different types of penalty is not stressed. 
The choice of $q$ and $\lambda$ are tuned by cross-validation, which cannot be implemented in many applications. 
Moreover, their approach was focused on deriving a generic class of regularized algorithms. 
While we find this work intriguing, our focus on applications of forecasting  and requires a more pragmatic approach to the accuracy criteria.

\section{Numerical Experiments}
\label{sec:exp}
 \begin{figure}[htp]
    \centering
    \begin{minipage}{.33\textwidth}
        \centering
        \includegraphics[width=\textwidth]{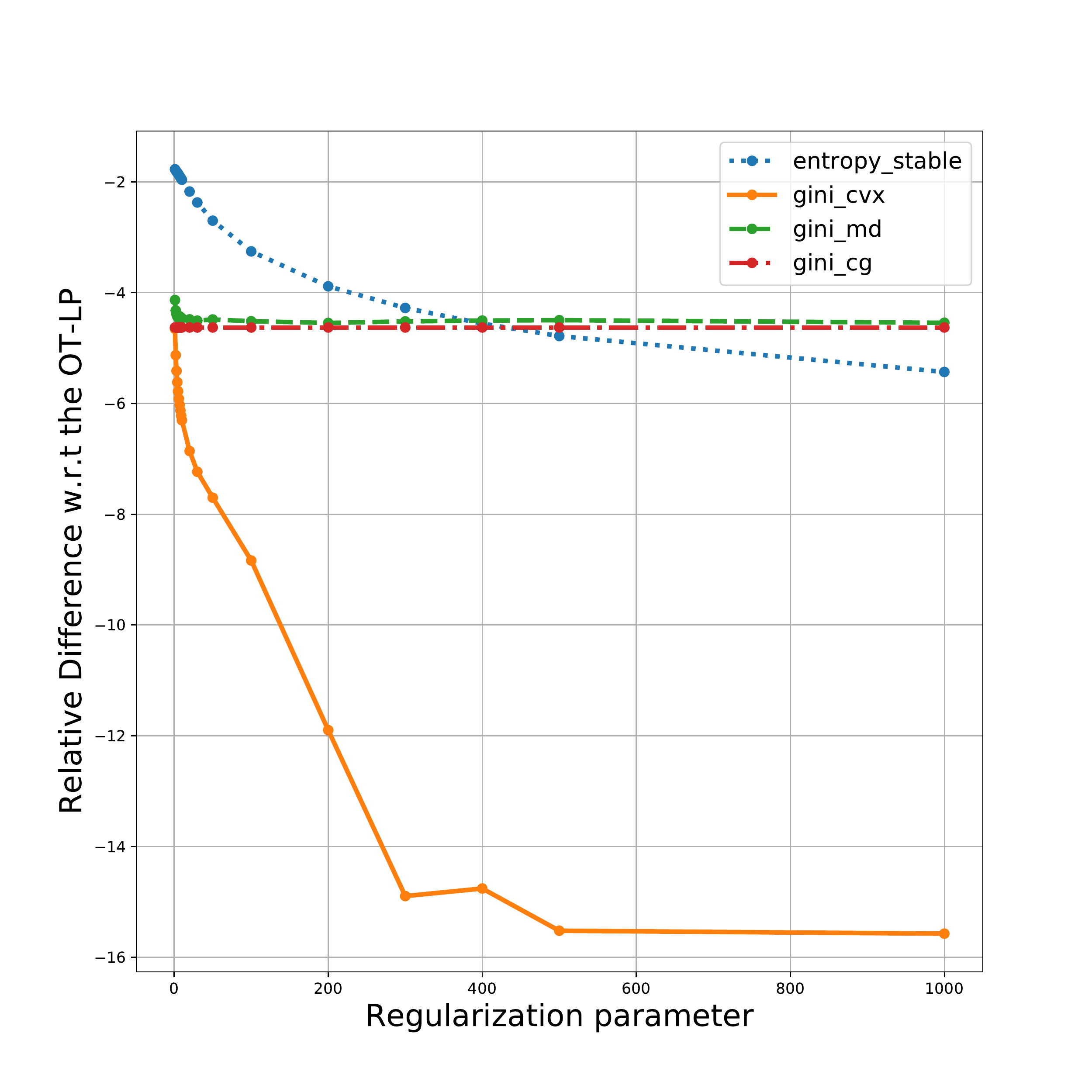}        
    \end{minipage}%
    \begin{minipage}{.33\textwidth}
        \centering
        \includegraphics[width=\textwidth]{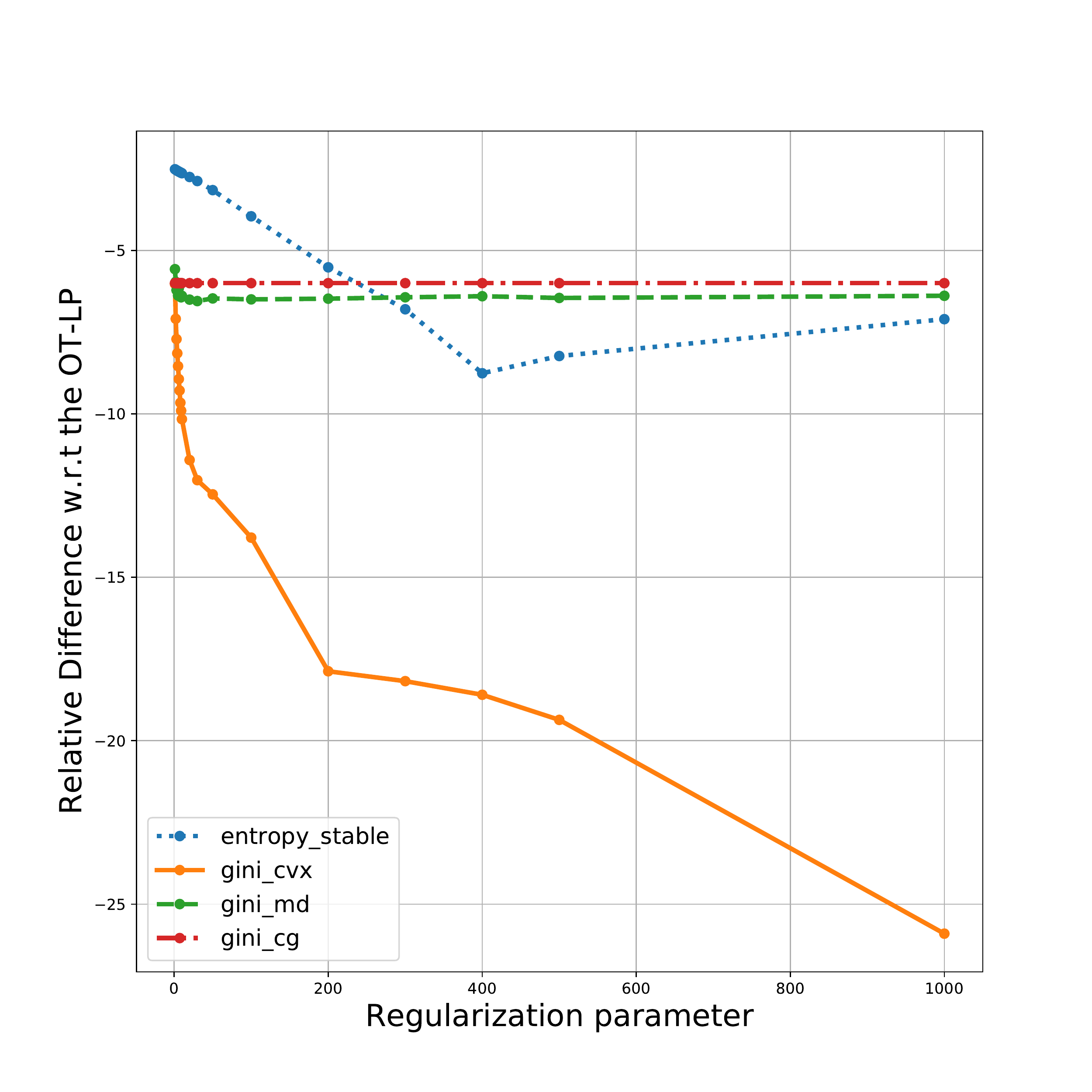}
    \end{minipage}
    \begin{minipage}{.33\textwidth}
        \centering
        \includegraphics[width=\textwidth]{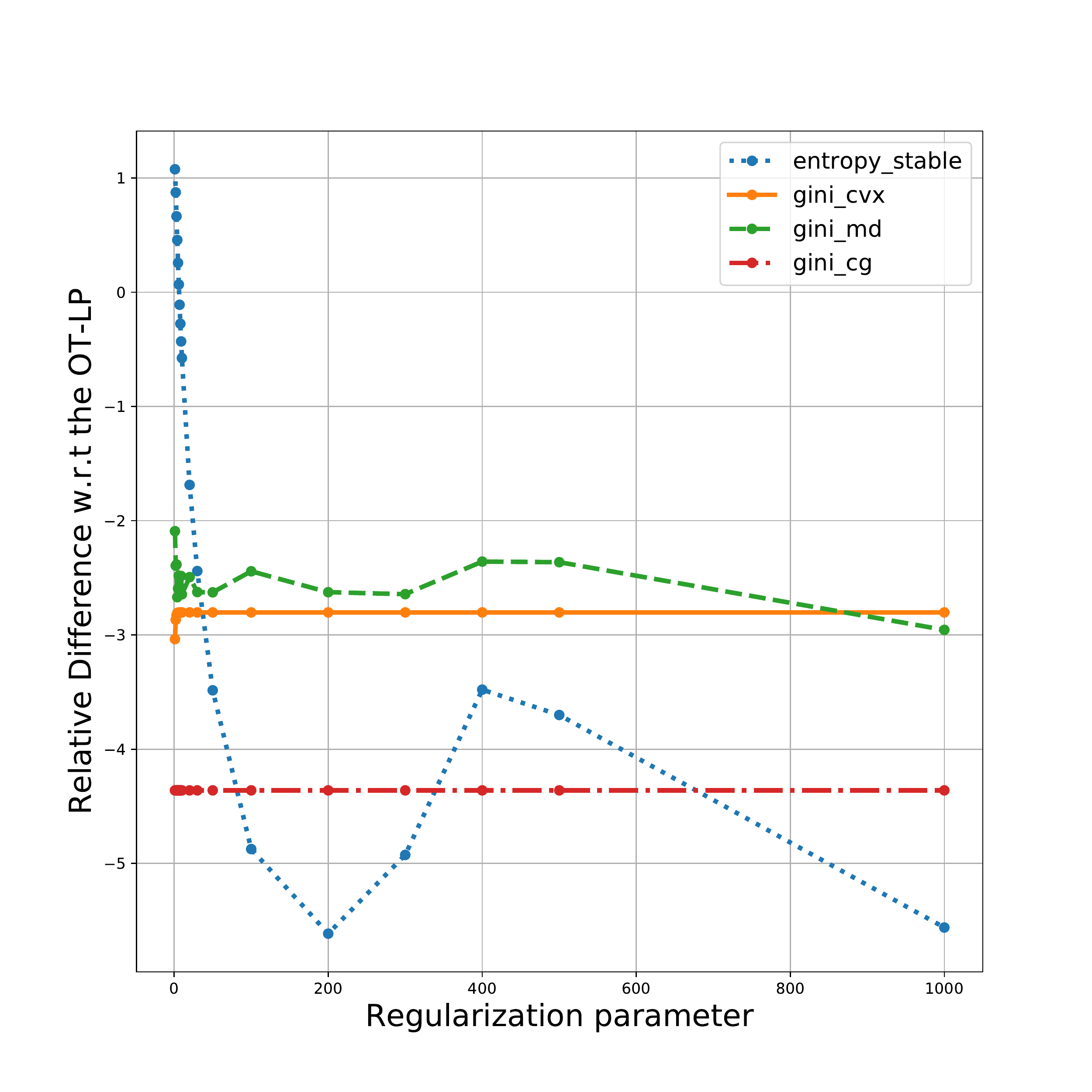}      
    \end{minipage}%
    \caption{Relative difference with respect to the OT-LP (in log scale) vs. $\lambda$ for (left) 1D Gaussian; (center) normalized random uniform; (right) logit-normal. }
    \label{fig:synthetic}
\end{figure}

\emph{Synthetic examples.} Figure~\ref{fig:synthetic} shows the effect of the regularization parameters with different synthetic datasets. We compare the stabilized Sinkhorn algorithm~\citep{schmitzer2016stabilized} with Gini-regularized OT with 3 solvers, namely, \textsc{cvx}, mirror descent (MD) and the conditional gradient (CG)~\citep{ferradans2014regularized} method. We use the PythonOT package~\citep{flamary2017pot} to calculate the classical OT solution and solve \eqref{eqn:got} with CG. The synthetic data, with three different underlying distributions, are generated as follows:

\begin{enumerate}
\item Wasserstein distance between $\mathcal{N}(20, 5)$ and $\mathcal{N}(50, 20)$ with $k=100.$
\item OT between two normalized uniform random with a scaled uniform random cost matrix with $k=200$. 
\item OT between two logit-normal densities. 
\end{enumerate}
While the normal and the uniform are likely familiar to the reader we briefly describe the logit-normal density. The logit-normal density is obtained by applying a logistic or soft-max transform to a random vector drawn from a multivariate normal density, resulting in a random vector of non-negative values whose sum is $1$. The logit-normal density allows us to approximate a Dirichlet distribution by setting the mean vector of the underlying normal density to the zero vector and the covariance matrix to be a small scalar multiple of the identity matrix. While the logit-normal can approximate the Dirichlet, the logit-normal can also model correlation structure on the unit simplex that is not achievable by the Dirichlet \citep{blei2006dynamic}. 

\begin{figure}[htp]
        \centering
        \includegraphics[width=0.5\textwidth]{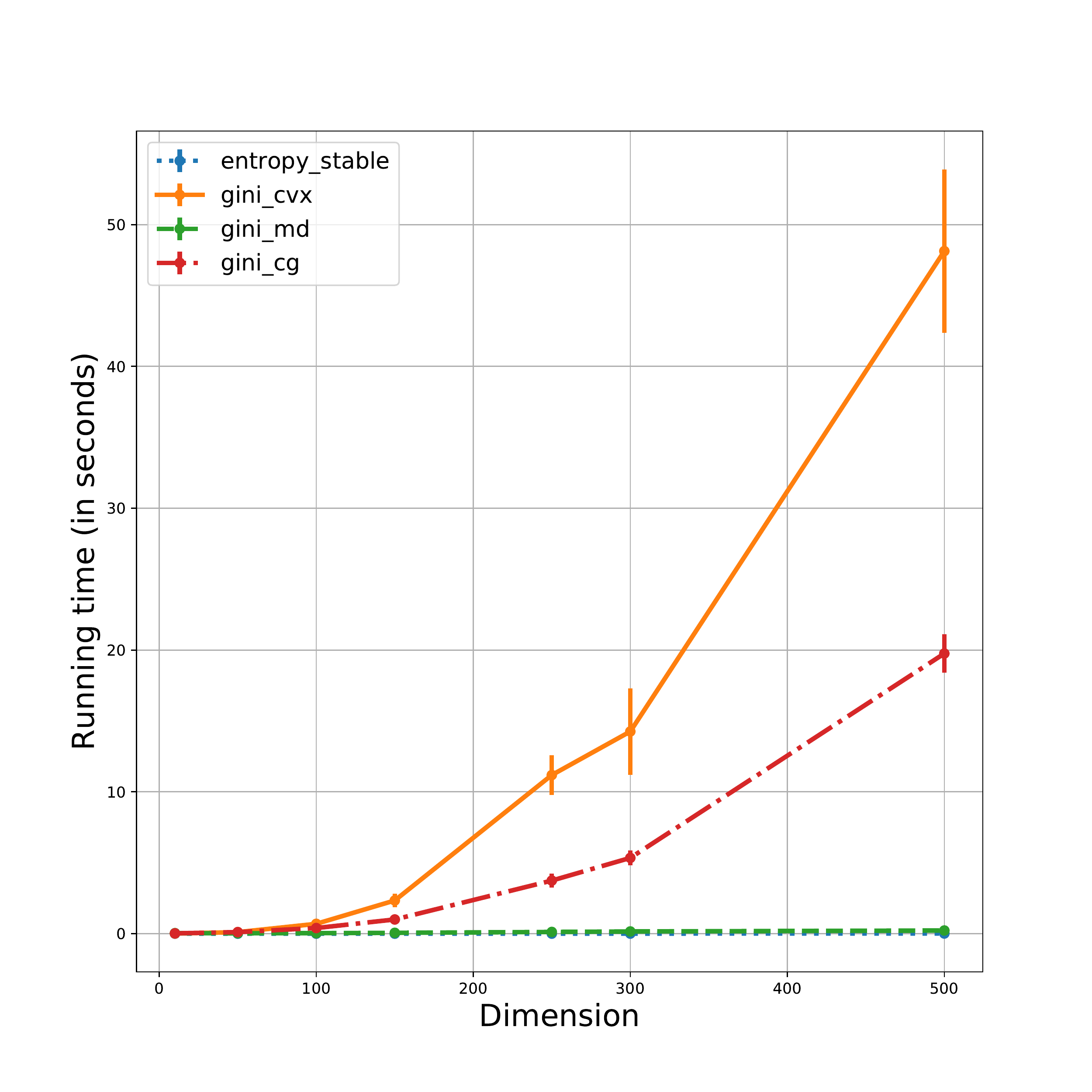}              
    \caption{Running time comparison vs. increasing dimensionality for synthetic data generated with model number 2. For each dimension, we repeat the experiments 10 times and report the errorbars. }
    \label{fig:exp3}
\end{figure}

Figure~\ref{fig:synthetic} shows that the Gini regularizer generally yields a more accurate solution when the regularization parameter is small (all 3 panels). 
As $\lambda$ increases, the \text{cvx} solver continues to improve while the accuracies are nearly constant for the MD and CG methods. 
For two of the three datasets the stabilized Sinkhorn method exhibits the strange behavior of worse results when $\lambda$ passes a threshold. 
Figure~\ref{fig:exp3} illustrates the running time (in seconds) vs. the dimensionality of the simplex. 
As we expected, \textsc{cvx} provides the most accurate solution but is expensive; CG scales better than \textsc{cvx} while the simple approach of MD shows promising scalability performance. 

It is worth noting that even for the stabilized Sinkhorn algorithm, one needs to manually choose the regularization parameter value through trial and error to avoid overflow/underflow errors. However, for the Gini regularizer, one is free from parameter tuning while maintaining relatively accurate approximations. 

\emph{Evaluating \regional\ forecasts.} Typically a \regional\ forecast would be evaluated during each time period. 

While large scale evaluations and concomitant technical issues of scaling are beyond the scope of this manuscript, we provide a single simulated product  example to give the reader a flavor of the application. 
We consider evaluating the \regional\ forecast at the state-level for illustration purposes. 

\begin{figure}[htp]
    \centering
    \begin{minipage}{.33\textwidth}
        \centering
        \includegraphics[width=\textwidth]{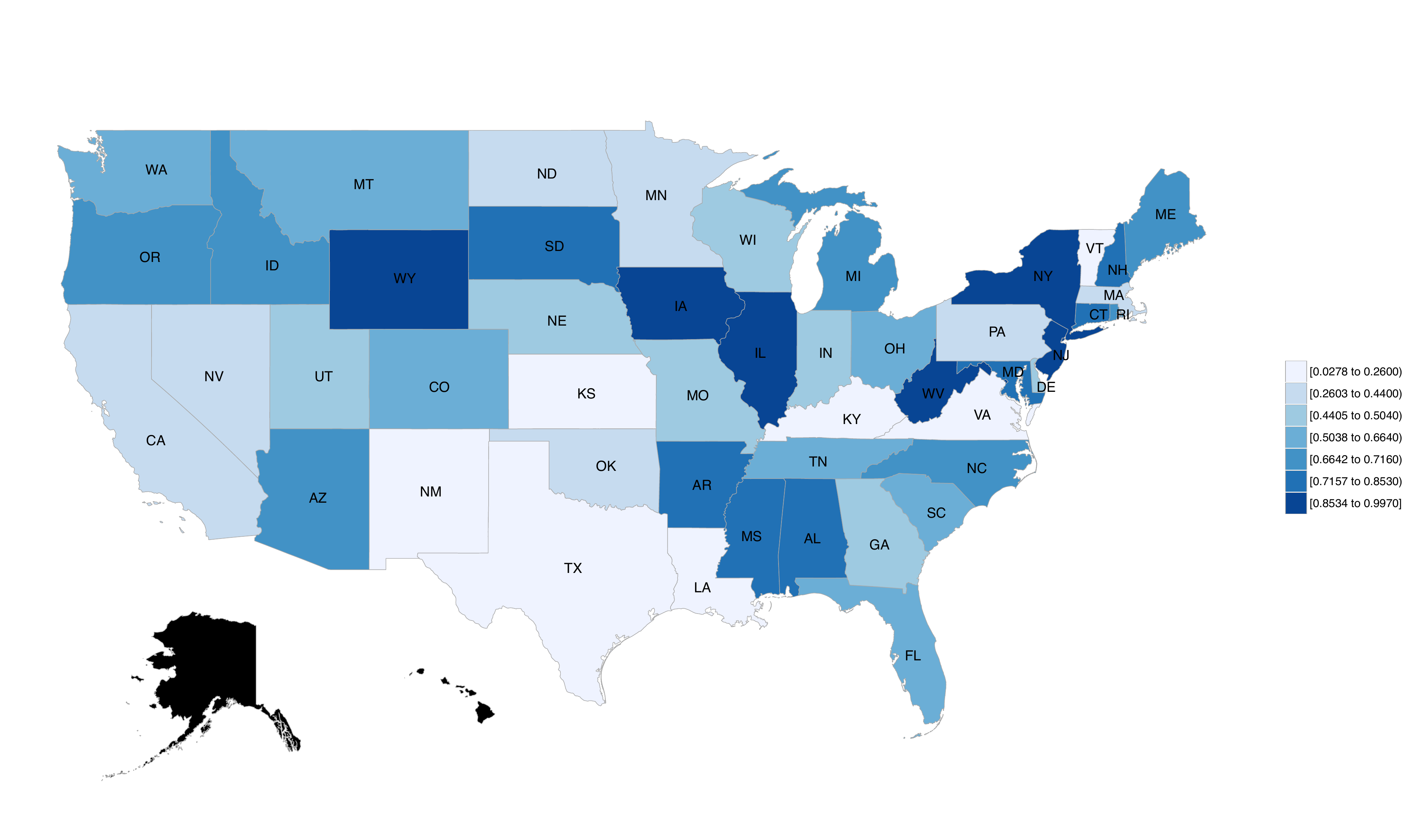}        
    \end{minipage}%
    \begin{minipage}{.33\textwidth}
        \centering
        \includegraphics[width=\textwidth]{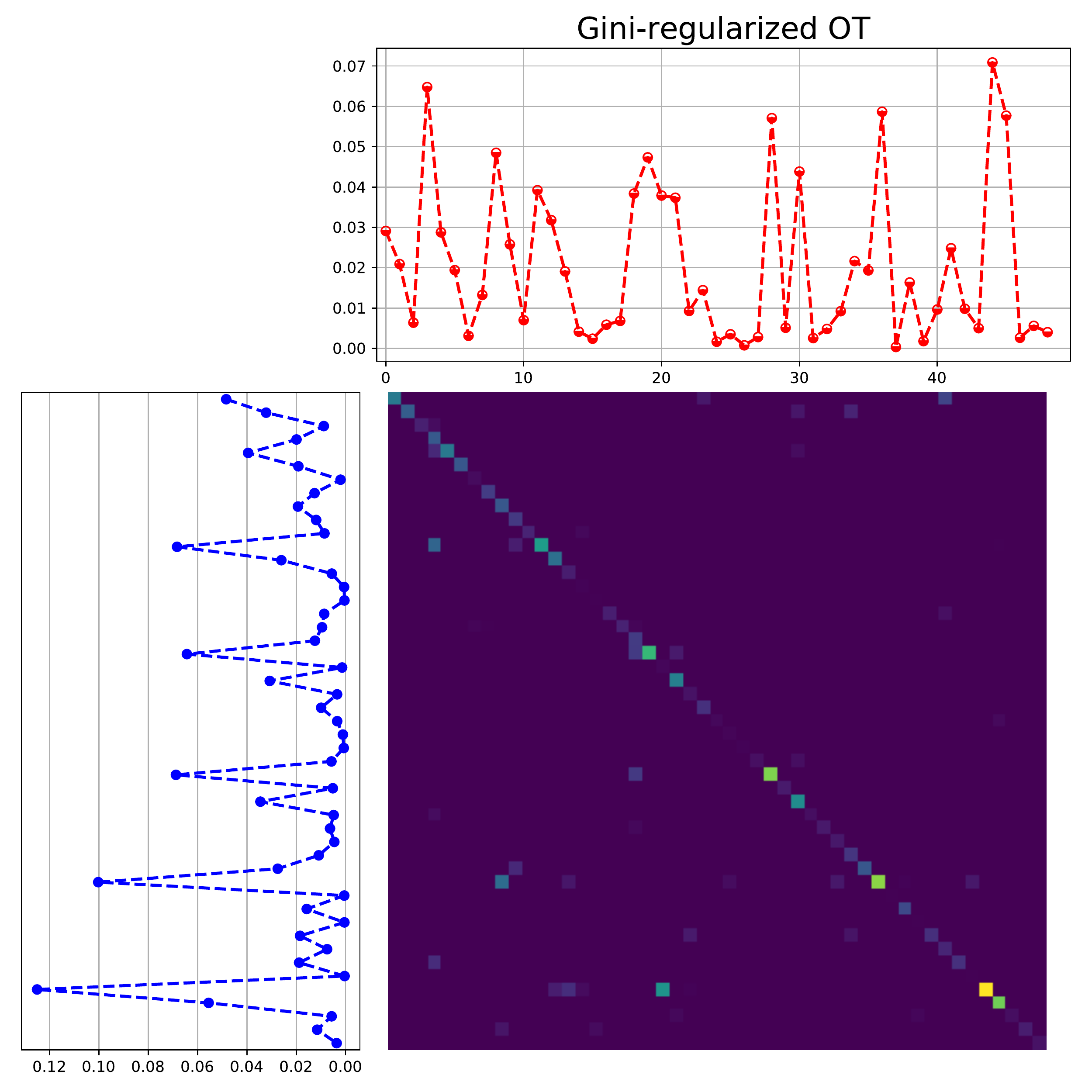}        
    \end{minipage}%
    \begin{minipage}{.33\textwidth}
        \centering
        \includegraphics[width=\textwidth]{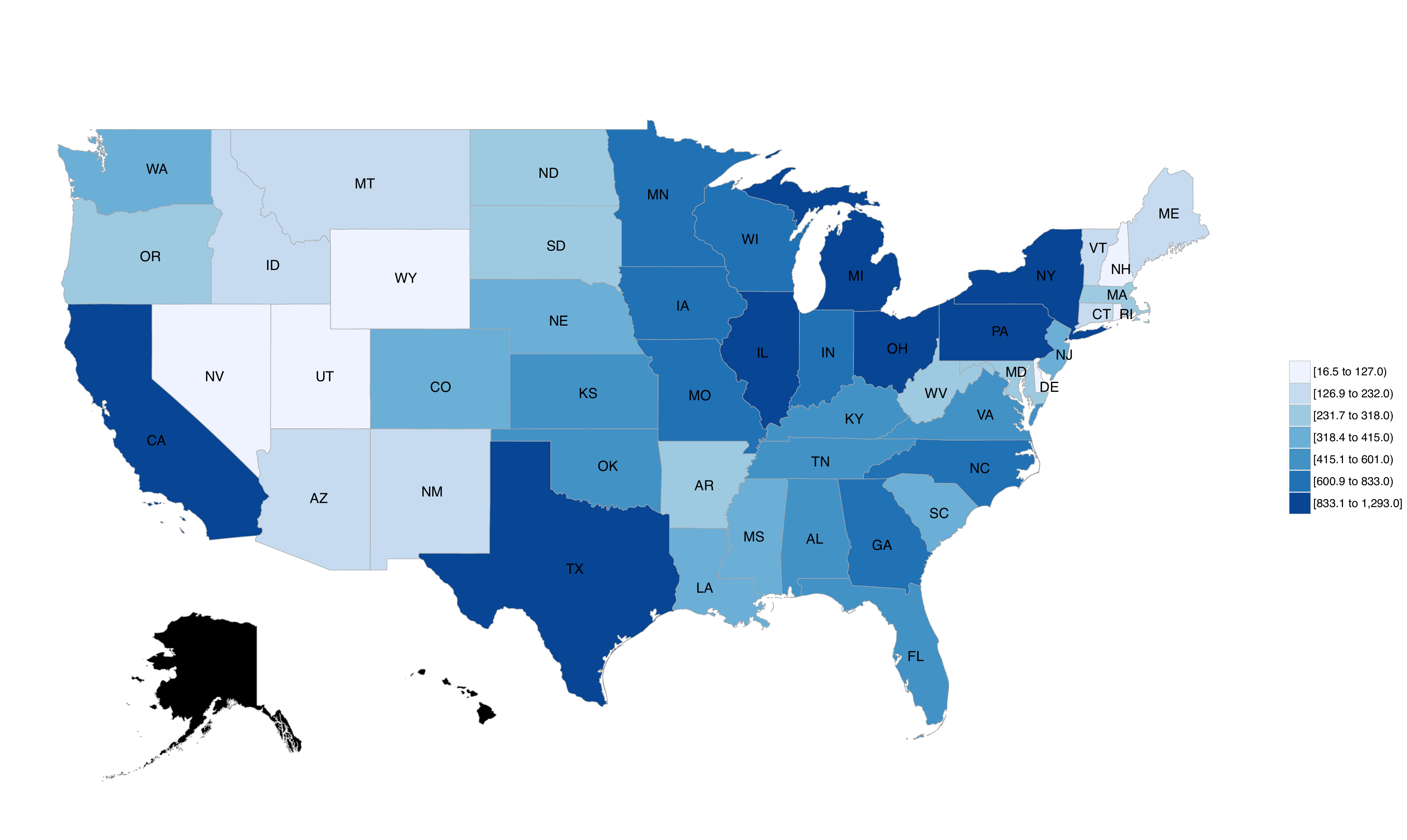}        
    \end{minipage}%
    \caption{An example of \regional\ forecast evaluation. Left: \regional\ forecast; Right: simulated sales; Middle: optimal transportation plan based on the geodesic distance.}
    \label{fig:exp2}
\end{figure}

The source measure is the forecasted distribution of demand for the products while the target measure is
the \todo{Leo: replace with 'simulated sales'} simulated sales for the product. 
The transport plan displayed in the center panel of Figure \ref{fig:exp2} shows the OT plan
to move the \regional\ forecasts to the simulated sales.

\section{Conclusions and Future Work}\label{sec:conc}
Based on OT theory we develop a forecasting accuracy criteria to evaluate spatio-temporal forecasts that account for the configuration of the distribution network. 
To achieve an accurate OT distance efficiently, we regularize the OT problem using the Gini impurity function.
 Unlike the entropic regularizer, the Gini regularizer is robust to changes of $\lambda$ and provides a more accurate approximation even when $\lambda$ is small. 
 This removes the need to manually tune $\lambda$, which is important in practice. 
 Future work will include developing a \regional\ forecasting method that optimizes the proposed accuracy criteria and investigate faster algorithms to solve the Gini-regularized OT. 
\section*{Acknowledgment}
We appreciate early discussions with Zhihao Cen about using EMD as an evaluation metric for spatio-temporal forecasting. We would also like to thank Ping Xu for her gracious support during the preparation of the manuscript. 

\bibliographystyle{plainnat}   
\bibliography{impurity_regularizers} 

\end{document}